\documentclass{article}
\usepackage{iclr2026_conference,times}
\usepackage{amsmath,amsfonts,bm}

\def\eqref#1{equation~\ref{#1}}
\def\1{\bm{1}}

\DeclareMathAlphabet{\mathsfit}{\encodingdefault}{\sfdefault}{m}{sl}
\SetMathAlphabet{\mathsfit}{bold}{\encodingdefault}{\sfdefault}{bx}{n}

\usepackage{algorithm}
\usepackage{hyperref}
\usepackage{url}
\usepackage{graphicx}
\usepackage{subcaption}
\usepackage{booktabs} % For professional tables
\usepackage{wrapfig}  % For figures wrapped by text
\usepackage{amssymb,mathtools}
\usepackage[T1]{fontenc}
\usepackage[utf8]{inputenc}
\usepackage{textcomp}
\usepackage{verbatim}
\usepackage{enumitem}

\usepackage{float}

\title{Bilinear representation mitigates reversal curse and enables consistent model editing}

\author{Dong-Kyum Kim$^{1}$ \quad Minsung Kim$^2$ \quad Jea Kwon$^1$ \quad Nakyeong Yang$^{1,2}$ \quad Meeyoung Cha$^{1,3}$\thanks{Corresponding author. Email: mia.cha@mpi-sp.org}\\
$^{1}$MPI-SP, Germany \quad $^{2}$SNU, South Korea \quad $^3$KAIST, South Korea
}

\iclrfinalcopy
\begin{document}

\maketitle

\begin{abstract}
% The reversal curse---a language model's (LM) inability to infer an unseen fact ``B is A'' from a learned fact ``A is B''---is widely considered a fundamental limitation. We show that this is not an inherent failure but an artifact of how models encode knowledge. By training LMs from scratch on a synthetic dataset of relational knowledge graphs, we demonstrate that bilinear relational structure emerges in their hidden representations. This structure substantially alleviates the reversal curse, enabling LMs to infer unseen reverse facts. Crucially, we also find that this bilinear structure plays a key role in consistent model editing. When a fact is updated in a LM with this structure, the edit correctly propagates to its reverse and other logically dependent facts. In contrast, models lacking this representation not only suffer from the reversal curse but also fail to generalize edits, further introducing logical inconsistencies. Our results establish that training on a relational knowledge dataset induces the emergence of bilinear internal representations, which in turn enable LMs to behave in a logically consistent manner after editing. This implies that the success of model editing depends critically not just on editing algorithms but on the underlying representational geometry of the knowledge being modified.
The reversal curse---a language model's inability to infer an unseen fact ``B is A'' from a learned fact ``A is B''---is widely considered a fundamental limitation. We show that this is not an inherent failure but an artifact of how models encode knowledge. Our results demonstrate that training from scratch on synthetic relational knowledge graphs leads to the emergence of a bilinear relational structure within the models' hidden representations. This structure alleviates the reversal curse and facilitates inference of unseen reverse facts. Crucially, this bilinear geometry is foundational for consistent model editing: updates to a single fact propagate correctly to its reverse and logically dependent relations. In contrast, models lacking this representation suffer from the reversal curse and fail to generalize model edits, leading to logical inconsistencies. Our results establish that training on a relational knowledge dataset induces the emergence of bilinear internal representations, which in turn support language models in behaving in a logically consistent manner after editing. This suggests that the efficacy of language model editing depends not only on the choice of algorithm but on the underlying representational geometry of the knowledge itself.
\end{abstract}

\section{Introduction}
\label{sec:introduction}

Language models have become powerful tools for knowledge-intensive tasks, yet their reasoning capabilities often fall short of human-level logical consistency~\citep{berglund2024the, allen-zhu2025physics}. A prominent example is the \textit{reversal curse}: a model trained on ``A is the parent of B'' frequently fails to infer the reverse fact, ``B is the child of A.'' This failure suggests that models learn shallow, directional associations rather than robust, symmetrical relationships required for reliable inference. This limitation is particularly acute in \textit{model editing}, which seeks to update factual knowledge in a trained model without costly retraining from scratch~\citep{de-cao2021editing,meng2022locating}. An ideal edit must be logically coherent; an update to the primary relation (for example, ``A is the spouse of C'' to ``A is the spouse of D'') should naturally propagate to its logical corollaries (``D is the spouse of A'' and ``B is the child of D'').

%, yet current architectures fail to maintain this integrity.

% However, existing approaches struggle with this logical generalization. Model editing methods often fail to propagate updates to the entailed facts, requiring that both directions of a relationship be explicitly co-edited to avoid the reversal curse~\citep{thibodeau2022but,yao2023editing,hase2024fundamental}. This limitation raises a critical question: Are these failures an inherent flaw in the Transformer architecture, or are they an artifact of how models learn to represent knowledge? While prior efforts have focused on explaining the directional nature of autoregressive objectives~\citep{zhu2024towards, kitouni2024the}, the latter question—whether the geometry of internal knowledge representations is responsible—remains unexplored.

However, existing architectures struggle with such logical generalization. Current model editing methods fail to propagate updates to entailed facts, requiring the explicit co-editing of both directions of a relationship to avoid the reversal curse~\citep{thibodeau2022but,yao2023editing,hase2024fundamental}. This limitation raises a critical question: are these failures an inherent flaw in the Transformer architecture, or are they an artifact of how models represent knowledge? While prior efforts have focused on the limitations of autoregressive objectives~\citep{zhu2024towards, kitouni2024the}, emerging evidence indicates that these reasoning failures are tied to the specific geometry of learned representations. Language models have been shown to learn meaningful geometric structures for features like space and time~\citep{gurnee2024language, engels2025not}. Furthermore, disrupting this underlying topology during editing correlates with failures in reasoning~\citep{nishi2025representation}. While these studies highlight that structure matters, the specific algebraic mechanism required to support symmetric and compositional reasoning remains an open question.

\begin{figure}[t]
    \centering
    \includegraphics[width=\textwidth]{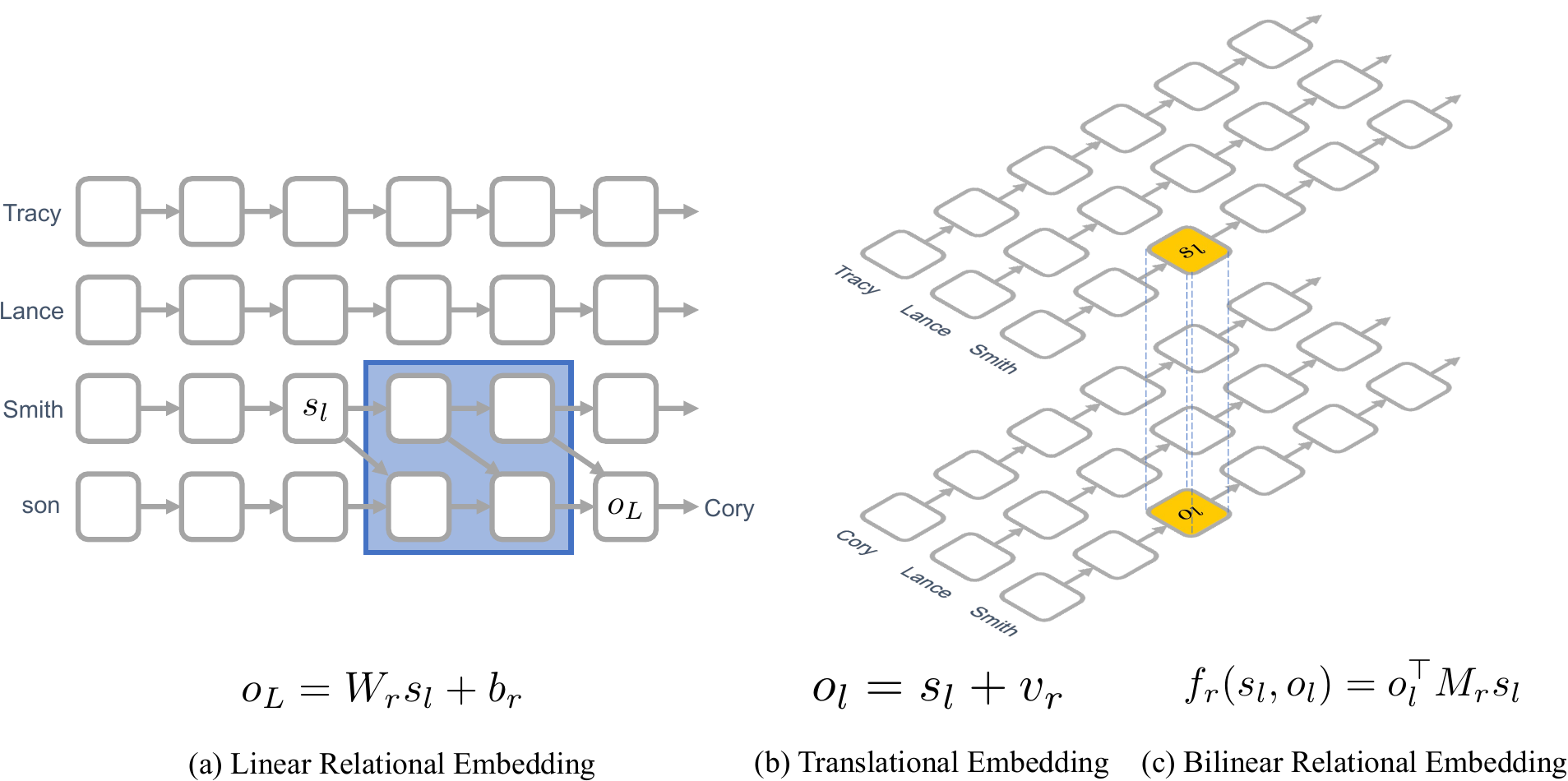}
    \caption{Schematics of relational embedding structures. For a subject $s$ and object $o$, the relation $r$ is modeled as: \textbf{(a) linear transformation, (b) vector translation,} or \textbf{(c) bilinear interaction} mediated by a relation-specific matrix $M_r$. As an example, in the fact ``Son of Tracy Lance Smith is Cory Lance Smith,'' $s$ is ``Tracy Lance Smith,” $o$ is ``Cory Lance Smith,'' and $r$ is son.}
    \label{fig:embedding_schematics}
\end{figure}

This work proposes that logical consistency in language models depends on the emergence of a \textit{bilinear} relational structure. We investigate how relational knowledge is encoded using three representation probes: linear, translational, and bilinear probes (see Figure~\ref{fig:embedding_schematics}). Recent work has demonstrated that knowledge decoding from a pretrained language model's internal representation can be approximated by either linear mapping~\citep{hernandez2024linearity} or translational mapping~\citep{merullo2024language}. While these frameworks offer valuable insights, they cannot capture symmetric and compositional relations essential for robust logical reasoning. We instead focus on bilinear relational models such as RESCAL~\citep{nickel2011three} in knowledge graph embedding literature, which represent relations as matrices that mediate interactions between entities. Bilinear relational structures naturally accommodate inverse relations via matrix transposition and compose relations via matrix multiplication, providing a rich algebraic framework for reasoning.

To investigate this, we train decoder-only Transformers from scratch on a synthetic knowledge graph, allowing us to precisely control the learning environment and probe the resulting internal structures. We make several contributions:
\begin{itemize}
    \item \textbf{Mitigate the reversal curse via relational knowledge learning:} We demonstrate that under appropriate regularization (weight decay), language models overcome the reversal curse by learning a robust bilinear relational structure, achieving near-perfect accuracy on unseen reverse relations (Figure~\ref{fig:weight_decay_effect}, right).
    \item \textbf{Identify bilinear structure in hidden representations:} 
    %Using three different probes on hidden representations, we find that a bilinear probe best explains them, with the signal emerging in intermediate layers (Figure~\ref{fig:probe_accuracy}). 
    Using three distinct probes, we find that a bilinear probe best explains hidden representations, with the highest precision emerging in the middle layers (Figure~\ref{fig:probe_accuracy}).
    Furthermore, the learned relation matrices satisfy formal algebraic tests for composition and inversion (Figure~\ref{fig:relational_algebra}), validating the presence of a functional bilinear structure in language models.
    \item \textbf{Link representational geometry to model editing:} We establish a mechanistic link between representational geometry and editing consistency. Models possessing a bilinear structure successfully propagate edits to logically related facts, whereas those lacking it fail to generalize, even when the direct edit is successful (Figure~\ref{fig:edit_generalization}).
\end{itemize}

Building on these findings, our study introduces a substantial change in perspective: the key to resolving logical failures lies not solely in the model architecture or editing algorithm, but fundamentally in the \textit{geometric structure of learned knowledge representations}. This result suggests the pivotal role of relational encoding in shaping the reliability and robustness of model behavior.

\section{Related work}
\label{sec:related_work}

\paragraph{Reversal Curse.}
The reversal curse---the failure to infer ``B is A'' from ``A is B''---has been identified as a fundamental limitation of language models~\citep{berglund2024the}. Prevailing hypotheses attribute this to the directional nature of the autoregressive training objective, which preferentially models $P(\text{B}|\text{A})$ but not $P(\text{A}|\text{B})$ \citep{allen-zhu2025physics, zhu2024towards, kitouni2024the}. The nature of this failure is nuanced; \citet{lin2024delving} suggest that it may be an issue of knowledge retrieval rather than storage, as models that fail open-ended generation can succeed on multiple-choice questions where the answer is present in a prompt. Proposed solutions often directly target the training process. These include data augmentation via subword-level reordering in a sentence \citep{golovneva2024reverse}, generating reversed examples by language models \citep{lampinen2025generalization}, or modifying the training objective to be direction-agnostic \citep{kitouni2024the}. Although these studies offer practical mitigation, they often treat it as an unavoidable artifact of the training objective. Our work offers a different perspective by focusing on the model's internal representations.

% \paragraph{Model Editing and Logical Generalization.} 
% Model editing aims to update facts in LMs without costly retraining~\citep{de-cao2021editing}. Various model editing methods have been proposed, such as ROME~\citep{meng2022locating} and MEMIT~\citep{meng2023massediting}, which show better generalization than naive fine-tuning, but all editing algorithms suffer from the reversal curse~\citep{thibodeau2022but, yao2023editing}. This points to a deeper conceptual challenge. \citet{hase2024fundamental} provide a comprehensive critique, framing model editing as an instance of belief revision, for which no simple solution exists. They argue that logical generalization is not just a desirable feature but a core requirement for any successful editing paradigm, and they demonstrate empirically that current methods often fail to produce coherent belief updates. While these works call for better testbeds, the underlying mechanisms causing these failures remain an open question. In our work, instead of proposing new editing algorithms, we investigate the prerequisite internal structures that enable logical consistency after model editing.

\paragraph{Model Editing and Logical Generalization.} 
Model editing aims to update facts in language models without costly retraining~\citep{de-cao2021editing}. Various model editing methods have been proposed, such as ROME~\citep{meng2022locating} and MEMIT~\citep{meng2023massediting}, which show better generalization than naive fine-tuning, but all editing algorithms suffer from the reversal curse~\citep{thibodeau2022but, yao2023editing}. This points to a deeper conceptual challenge. \citet{hase2024fundamental} provide a comprehensive critique, framing model editing as an instance of belief revision, for which no simple solution exists. They argue that logical generalization is not just a desirable feature but a core requirement for any successful editing paradigm, and they demonstrate empirically that current methods often fail to produce coherent belief updates. Recently, \cite{nishi2025representation} demonstrate that editing can break the underlying topological structure of the learned knowledge. This suggests that the brittleness of model editing is tied to the preservation of internal geometry. In our work, we investigate the prerequisite internal algebraic structures that enable logical consistency after model editing.

\paragraph{Mathematical Structures for Relational Knowledge in Language Models.} 
A substantial body of research has investigated \textit{where} language models store factual knowledge~\citep{geva2021key,meng2022locating,pan2025precise}. The mechanism for retrieving specific facts is complex and distributed across multiple layers and attention heads, as revealed by studies of interventions~\citep{hase2023does} and attention mechanisms~\citep{geva2023dissecting}. In contrast, far fewer studies examine \textit{which mathematical structure} models employ to resolve a relation. Recent investigations suggest that the underlying structures are unexpectedly simple: \citet{hernandez2024linearity} demonstrate that language models implicitly implement Linear Relational Embeddings~\citep{paccanaro2002learning}, while \citet{merullo2024language} show that they exploit translational structures familiar from Word2Vec~\citep{mikolov2013word2vec}. However, the multi-head attention mechanism is fundamentally built on more expressive bilinear operations~\citep{elhage2021mathematical} and individual heads can encode specific relational operations~\citep{elhelo2025inferring}. Furthermore, such bilinear models have a long history of success in modeling relational data, particularly in knowledge graph embedding methods like RESCAL~\citep{nickel2011three}. Despite bilinear operations being central to Transformer architecture and successful in relational learning, little work has investigated whether language models exploit bilinear structures for decoding relational knowledge. This paper presents a systematic analysis of the bilinear structures underlying relational knowledge.

\section{Relational Knowledge Dataset and Language Models}
\label{sec:method}

In order to test our hypothesis, we create a synthetic relational knowledge dataset and train multiple language models with different hyperparameters from scratch on it. 

\begin{figure}[t]
    \centering
    \includegraphics[width=\textwidth]{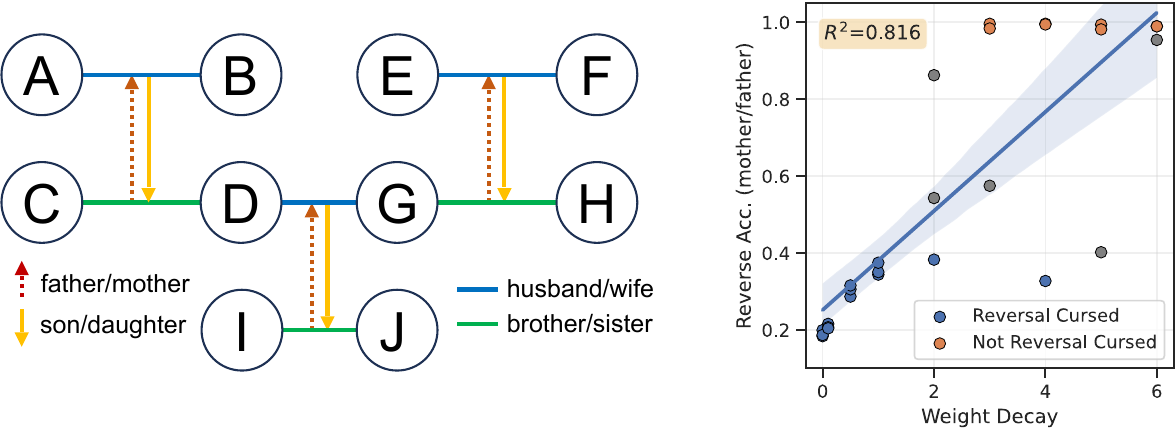}
    \caption{
        \textbf{(Left)} Schematic of the synthetic family knowledge graph. Nodes represent entities, while directed edges represent one of eight distinct family relations used in our experimental evaluation.
        \textbf{(Right)} Test accuracy on the unseen relations (\texttt{mother}/\texttt{father}) as a function of weight decay. Each weight decay setting was trained using three different random seeds. 
    }
    \label{fig:weight_decay_effect}
\end{figure}

\paragraph{Synthetic Knowledge Graph and Task.}
\label{subsec:data}
We generate a dataset from a synthetic family knowledge graph (see Figure~\ref{fig:weight_decay_effect}, left). The graph consists of entities (family members) and eight relations: \texttt{husband}, \texttt{wife}, \texttt{father}, \texttt{mother}, \texttt{son}, \texttt{daughter}, \texttt{brother}, and \texttt{sister}. We choose this domain because family relations form a minimal, closed-world system that exhibits inverses (\texttt{husband} of \texttt{wife} is \texttt{husband}) and composition/multi-hop structure (e.g., \texttt{husband} of \texttt{mother} is \texttt{father}, \texttt{sister} of \texttt{son} is \texttt{daughter}) under clear type constraints; the eight relations are the smallest set that jointly spans these algebraic properties, making the setup ideal for testing reversal and logical generalization.

Each entity is assigned a full name following the format ``[First Name] [Middle Name] [Last Name]''. All entities of a family share a common family name, defined as ``[Middle Name] [Last Name]''. In this work, we denote a fact as $(s, r, o)$ where $s$ is the subject entity, $r$ is the relation, and $o$ is the object entity; this means $r$ of $s$ is $o$. Each fact is represented as a plain text sentence: ``[Subject First Name] [Family Name] [Relation] [Object First Name] [Family Name]''. For example ``Emily Scott Wall husband Julian Scott Wall'' where ``Scott Wall'' is the family name.

The dataset comprises 1,000 families, each with 10 entities, resulting in 36 distinct relational facts per family. These families are divided into two groups of 500 to create the training set.
\begin{itemize}
\item \textbf{Group 1 (Full relations):} For the first 500 families, this group contains all 36 facts.
\item \textbf{Group 2 (Missing relations):} For the second 500 families, we withhold the \texttt{father} and \texttt{mother} relations. Therefore, this group contains only the remaining 24 facts.
\end{itemize}

The training set consists of all facts from both groups (approximately 318M tokens). 
The test set is constructed exclusively from the withheld facts of Group 2, containing 12 relations (\texttt{father}/\texttt{mother}) per family. 
The task is to predict these unseen relations in Group 2 by learning logical dependencies between entities from Group 1. 
If the model has learned a relational structure in the dataset, then it can infer these missing relations by using logical reasoning, e.g. the test relation (C, father, B) can be inferred by using two facts (A, husband, B) and (B, son, C). Full details of our synthetic dataset are provided in Appendix~\ref{appendix:data}.

\paragraph{Language Models.}
\label{subsec:model_training}
We train decoder-only Transformers using the GPT-NeoX architecture~\citep{gpt-neox-library} from scratch on the synthetic dataset. Each model has 12 layers, a hidden size of 896, and 16 attention heads, totaling approximately 206M parameters. We employ this architecture in the following sections. More details of architecture and training are provided in the Appendix~\ref{appendix:training_details}.

\section{Experiments}
\label{sec:experiments}

Our experiments seek to uncover the internal mechanisms that enable language models to perform logical reasoning. We first demonstrate that even when the training data contain sufficient information to infer reverse relations, models only overcome the reversal curse when guided by appropriate regularization (Experiment 1). We then pivot to the central question of our work: \textit{which mathematical structure} enables this success? Through a series of probing experiments, we uncover an emergent bilinear relational structure in the successful models (Experiment 2) and verify its algebraic properties (Experiment 3). Finally, we demonstrate that this geometry plays a central role in addressing model editing challenges. We show that the bilinear structure is a key to ensuring that edits propagate in a logically consistent manner, establishing a unified explanation for both the reversal curse and editing generalization failures (Experiment 4).

\subsection{Experiment 1: Training language models on relational knowledge dataset}
\label{subsec:reversal_curse}

We test whether a language model can learn to infer withheld relations (\texttt{father}/\texttt{mother}) by observing their reverse counterparts (e.g., \texttt{son}/\texttt{daughter}) and compositional examples elsewhere in the training set. We use AdamW~\citep{loshchilov2018decoupled} with a learning rate of $3 \times 10^{-4}$ and sweep weight decay over $\{0, 0.1, 0.5, 1.0, 2.0, 3.0, 4.0, 5.0, 6.0\}$, training three random seeds per setting (27 models in total). 

The results, shown in Figure~\ref{fig:weight_decay_effect} (right), reveal a striking dependency. All models achieve 100\% training accuracy (see Appendix~\ref{appendix:results} and Figure~\ref{sfig:train_loss}), but test accuracy varies significantly based on weight decay and random seed. Without sufficient regularization, models consistently fail, showing the reversal curse with low accuracy (weight decay$~<~$1.0). However, as weight decay is increased, a split in outcomes emerges: some models remain ``reversal cursed,'' while others break the curse and achieve near-perfect accuracy on the unseen reverse facts. This transition indicates that the reversal curse is not an inherent limitation but an artifact of an under-constrained training objective, where regularization promotes a more generalizable internal structure over simple memorization. 

This finding motivates our subsequent experiments. To understand the mechanisms distinguishing these two outcomes, we select two representative models for in-depth analysis: a ``Reversal Cursed'' model with low reverse accuracy ($<~$40\%) and a ``Not Reversal Cursed'' model with high reverse accuracy ($>~$98\%). In all subsequent figures, the former is indicated by blue, and the latter by orange.

\subsection{Experiment 2: Probing internal representation for relational structures}
\label{subsec:probing}

To identify which relational geometry the models have learned, we conduct a probing analysis. The process begins by extracting entity representations for each fact $(s, r, o)$. Specifically, we take the hidden states for subject ($s_l$) and object ($o_l$) from layer $l$ at the final token of their names, resulting in vectors in $\mathbb{R}^{d}$ where dimension $d=896$. Here, $r$ denotes a relation. Using these representations, we train three different probes to test our structural hypotheses (Figure~\ref{fig:embedding_schematics}). The training set consists of facts from 125 families (1{,}250 entities) from Group 1, and the evaluation is performed on a held set of facts from 125 different families, also from Group 1. The resulting classification accuracy indicates the degree to which a given relational structure is present at each layer.

\paragraph{Linear Relational Embedding.} 
This probe tests for a linear relational structure~\citep{paccanaro2002learning} in language models. \citet{hernandez2024linearity} model a relation $r$ as the local affine transformation that the Transformer applies to map a subject representation $s_l$ from layer $l$ to the object's pre-prediction representation $o_L$ in the final layer $L$. This yields the approximation $o_L \approx W_r s_l + b_r$, where $o_L$ is the hidden state at the position immediately preceding the object token (see Figure~\ref{fig:embedding_schematics}a).

The parameters $\{W_r, b_r\}$ are not learned but are extracted directly from the model's forward pass. The matrix $W_r$ is estimated from the Jacobian $J_r = \partial o_L / \partial s_l$, averaged over $n$ training examples. As the raw Jacobian can underestimate the transformation's magnitude, we scale it by a hyperparameter $\beta$, chosen via a sweep over $\{1.0, 1.5, 2.0, \dots, 5.0\}$. The bias $b_r$ is then computed as the mean residual. The parameters are thus set as:
\begin{equation}
W_r = \frac{\beta}{n}\sum_{i=1}^{n} J_r^{(i)}, \qquad b_r = \frac{1}{n}\sum_{i=1}^{n}\big(o_L^{(i)} - J_r^{(i)} s_l^{(i)}\big)    
\end{equation}
We estimate these parameters by averaging a small sample of $n=10$ training examples per relation. $n=10$ is chosen from a sweep over $\{10, 100, 500\}$ (See Appendix~\ref{appendix:lre_N}).

\paragraph{Translational.} This probe, which tests for the kind of vector arithmetic investigated in hidden representations by \citet{merullo2024language}, models a relation $r$ as a simple vector offset. For each relation $r$, we fit a translation vector $v_r$ such that $s_l + v_r \approx o_l$. Unlike the linear relation embedding, $o_l$ is taken from the same layer $l$ as $s_l$ (see Figure~\ref{fig:embedding_schematics}b) at the last token position of the object name.

The vector $v_r$ is computed as the average displacement across all $n$ training facts for that relation:
\begin{equation}
v_r = \frac{1}{n}\sum_{i=1}^{n}\big(o_l^{(i)} - s_l^{(i)}\big)
\end{equation}
This structure suggests that all entities participating in a relation are shifted by a constant vector in the embedding space.

\paragraph{Bilinear.}

This probe tests for a bilinear structure, where a relation is modeled as a matrix $M_r$ that mediates the interaction between the subject and object embeddings (see Figure \ref{fig:embedding_schematics}c). For each relation $r$, we define a score function: 
\begin{equation}
f_r(s_l, o_l) = s_l^\top M_r o_l,    
\end{equation}
between any pair of entities given the relation $r$ where matrix $M_r \in \mathbb{R}^{d \times d}$. The target function is:
\begin{equation}
f^*_r(s, o) = 
\begin{cases} 
1 & \text{when } (s, r, o) \text{ is true (exists in the dataset)}, \\
0 & \text{when } (s, r, o) \text{ is false (does not exist)}.
\end{cases}
\end{equation}
We estimate the relation matrices using a ridge regression variant of the RESCAL algorithm \citep{nickel2011three}, which optimizes the following objective function:
\begin{equation}
    L(M_r) = \frac{1}{2} \| \mathcal{X}_r - \mathbf{A} M_r \mathbf{A}^T \|_F^2 + \frac{\lambda_R}{2} \|M_r\|_F^2, \qquad \mathbf{A}^\top = [s^{(1)}_l, s^{(2)}_l, \ldots, s^{(n)}_l] \in \mathbb{R}^{d \times n},
\end{equation}
where $\mathbf{A} \in \mathbb{R}^{n \times d}$ is the matrix of trainset entity embeddings, $\mathcal{X}_r \in \mathbb{R}^{n \times n}$ is the adjacency matrix for relation $r$ (with entries corresponding to $f^*_r$), and $\lambda_R$ is a regularization parameter. Here, $n$ is the total number of entities ($n=1{,}250$). Due to the high dimensionality of our embeddings ($d=896$), we use an SVD-based optimization approach to make the computation tractable (detailed in Appendix \ref{appendix:rescal_svd}). 
We sweep the regularization parameter $\lambda_R$ over a logarithmic scale from $10^{-3}$ to $10^{-1}$ to find the optimal function for each layer $l$ and relation $r$.

\begin{figure}[t]
    \centering
    \includegraphics[width=\textwidth]{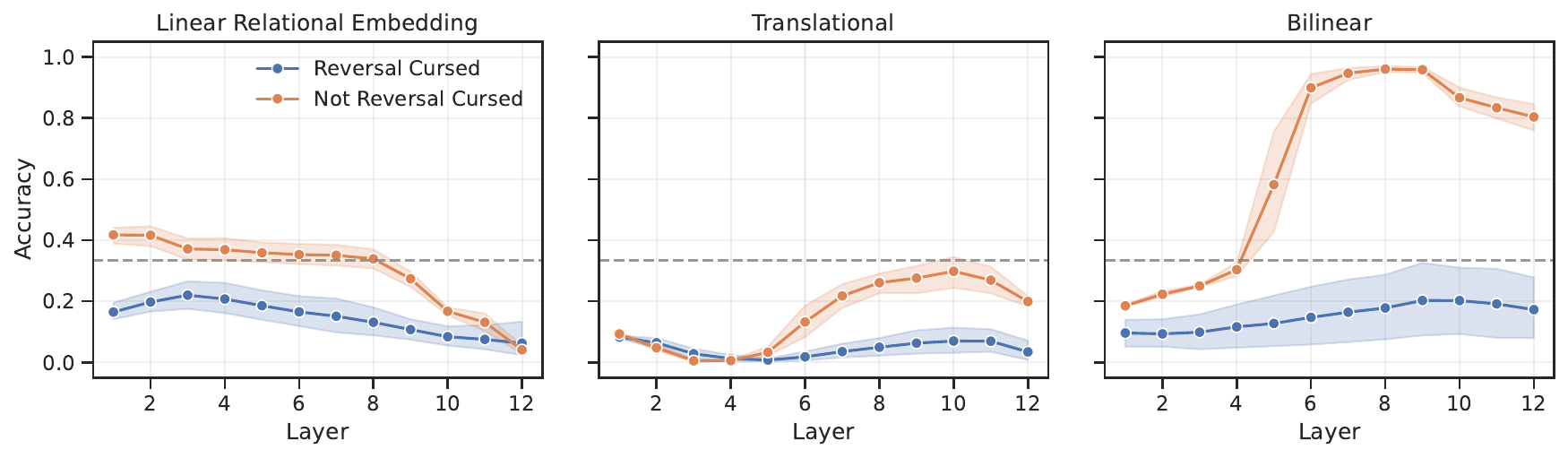}
    \caption{\textbf{Layer-wise probe accuracy for relational embeddings.} We compare ``Reversal Cursed'' models (blue) against ``Not Reversal Cursed'' models (orange), averaged across all relations. Accuracy is reported for each layer to identify where relational knowledge emerges.}
    \label{fig:probe_accuracy}
\end{figure}

\paragraph{Results.} For each layer $l$ and relation $r$, we trained a separate probe and evaluated its accuracy on test set entities (Figure~\ref{fig:probe_accuracy}). The gray dashed line at 1/3 marks the chance-level baseline. In our family graph (see Figure~\ref{fig:weight_decay_effect}, left), for any relation $r$ there are only three candidates per family that satisfy $f^*_r(s,o)=1$. A probe that identifies the family name and the relation $r$ but fails to use the subject embedding would therefore guess uniformly among these three candidates, yielding an expected accuracy of $1/3$. For example, given the text ``[Subject] Scott Wall husband [Object]'', a probe that ignores the subject can narrow the object to the three husband candidates in the ``Scott Wall'' family and would be correct one out of three times on average.

Against this baseline, the ``Not Reversal Cursed'' model (orange) develops a strong localized bilinear structure: the precision of the bilinear probe rises sharply in the middle layers (6--9), peaking above 95\%, while the linear and translational probes hover near or below the baseline. In contrast, the ``Reversal Cursed'' model (blue) shows no coherent relational geometry; all probes remain low and often near the 1/3 line across layers. Per-relation results (see Appendix~\ref{appendix:results}) show: (1) translational structure is confined to the symmetric relation \texttt{husband/wife}; (2) some ``Reversal Cursed'' models exhibit weak, relation-isolated linear relational mappings; (3) only ``Not Reversal Cursed'' models express a consistent high-fidelity bilinear pattern across all eight relations. These results indicate that the emergence of a bilinear representation is a key mechanism that enables the model to overcome the reversal curse. 

\subsection{Experiment 3: Relational Algebra Tests}
\label{subsec:relational_algebra}

\begin{figure}[t]
    \centering
    \includegraphics[width=\textwidth]{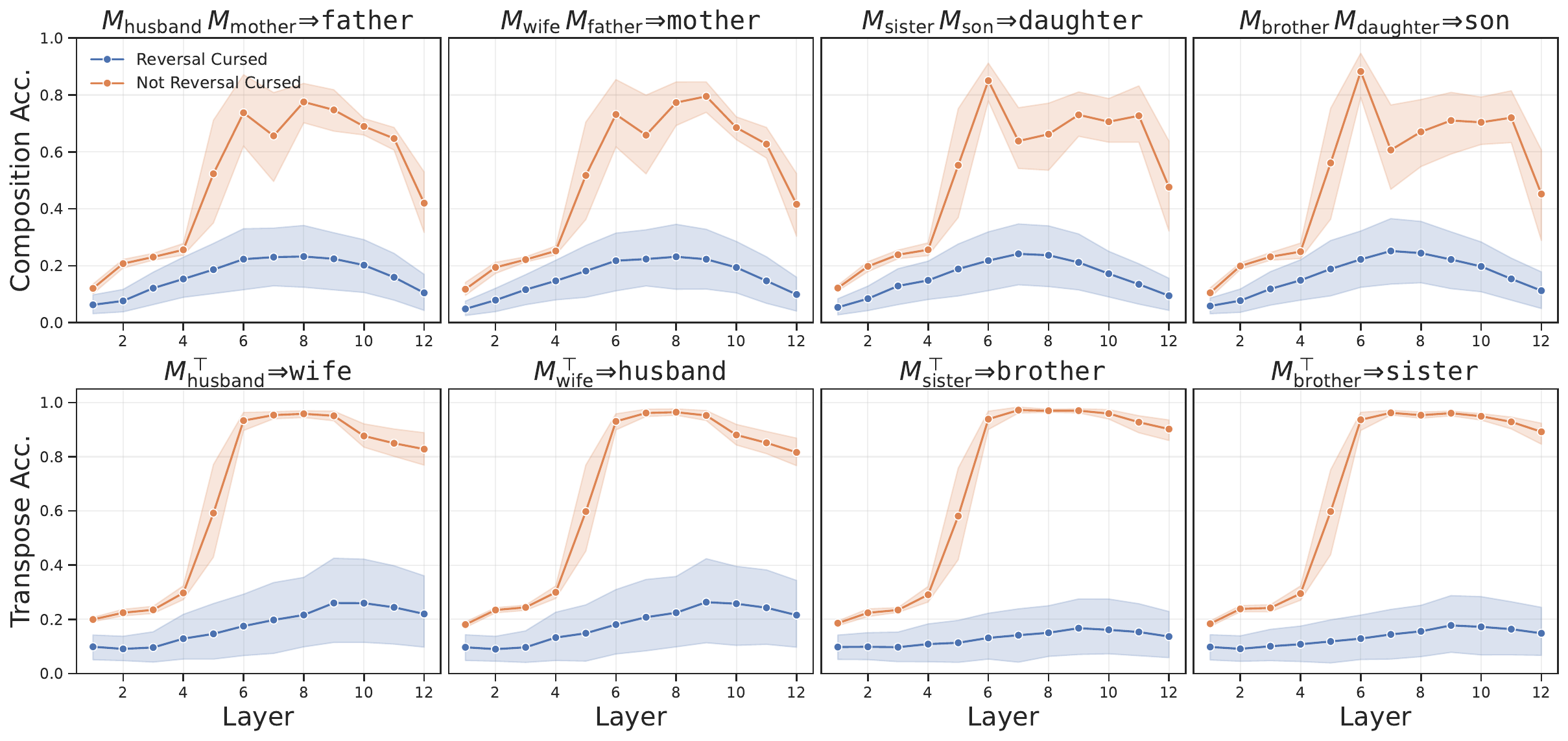}
    \caption{\textbf{Performance on formal algebraic tests.} \textbf{Top Row (Composition):} Accuracy of inferring a composed relation using the product of the corresponding bilinear matrices (e.g., $M_{\text{husband}} \cdot M_{\text{mother}}$ to probe for `father'). \textbf{Bottom Row (Transpose or Inversion):} Accuracy of inferring an inverse relation using the transpose of a bilinear matrix (e.g., $M_{\text{husband}}^\top$ to probe for `wife').}
    \label{fig:relational_algebra}
\end{figure}

Having established a strong bilinear signal in Experiment 2, we test whether the trained relation matrices $M_r$ obey basic algebraic laws that enable inverse and multi-hop reasoning. Using the matrices estimated by the bilinear probe in each layer, we evaluate two relational reasoning operations with the same scoring function $f_r$ and evaluation protocol as the bilinear probe: (1) \textbf{Transpose or Inversion}: does $M_{r}^\top$ act like the inverse relation $r^{-1}$? (2) \textbf{Composition}: does the product $M_{r_2} M_{r_1}$ act like the composed relation $r_2\!\circ\!r_1$?

We instantiate four compositions matching our dataset (Figure~\ref{fig:weight_decay_effect}, left):
$M_{\text{husband}} M_{\text{mother}}\!\Rightarrow\! \texttt{father}$,
$M_{\text{wife}} M_{\text{father}}\!\Rightarrow\! \texttt{mother}$,
$M_{\text{sister}} M_{\text{son}}\!\Rightarrow\! \texttt{daughter}$,
$M_{\text{brother}} M_{\text{daughter}}\!\Rightarrow\! \texttt{son}$.
For inversion, we test the pairs
$M_{\text{husband}}^\top\!\Rightarrow\!\texttt{wife}$,
$M_{\text{wife}}^\top\!\Rightarrow\!\texttt{husband}$,
$M_{\text{sister}}^\top\!\Rightarrow\!\texttt{brother}$,
$M_{\text{brother}}^\top\!\Rightarrow\!\texttt{sister}$.

\paragraph{Results.} Figure~\ref{fig:relational_algebra} shows that the ``Not Reversal Cursed'' model (orange) achieves high accuracy in both composition and transpose tests, with peaks aligned to the same middle layers (6--9) where the bilinear probe is strongest. The ``Reversal Cursed'' model (blue) remains low in all layers. These results indicate that the learned bilinear representation is not merely predictive but algebraically structured: transposes approximate inverse relations and matrix products approximate composed relations, enabling multi-hop inference.

\subsection{Experiment 4: Model editing and its link to bilinear structure}

\paragraph{Model editing and evaluation.} 
We edit a husband-relation fact (A, husband, B) and evaluate its effect on entailed knowledge. We conducted 50 editing experiments per model, each editing a single fact (A, husband, B) from Group 1 to (A, husband, B'), where B' has the same family name but a different first name than B. We perform edits using a straightforward yet effective layer-wise fine-tuning that minimizes cross-entropy loss on the new fact~\citep{zhu2020modifying,wang2024easyedit}. We use the Adam optimizer with a learning rate of $4 \times 10^{-4}$ and apply early stopping once the loss drops below 0.2 and restrict gradient updates to the MLP block's output layer. We apply this to each layer $l \in \{1, \ldots, 12\}$, yielding 12 edited models per original model.

\begin{figure}[t]
    \centering
    \includegraphics[width=\textwidth]{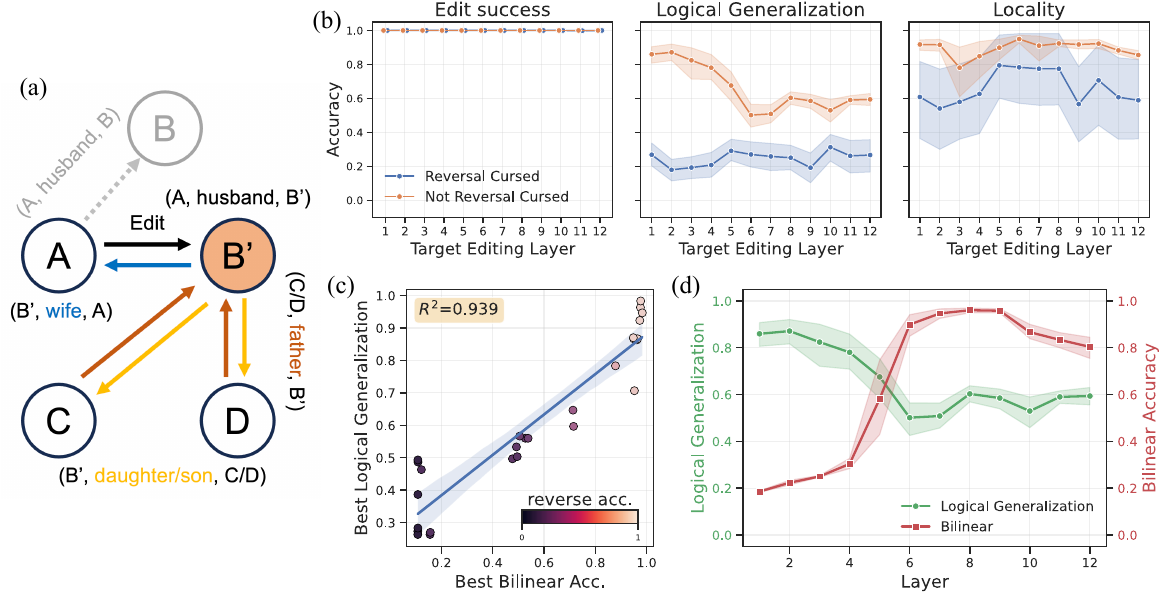}
    \caption{\textbf{Editing generalization and its link to bilinear structure.} \textbf{(a)} Schematic of the editing task. The primary fact (A, husband, B) is edited to (A, husband, B'). Successful logical generalization requires updating the inverse (B', wife, A) and neighborhood relations (C/D, father, B'); (B', daughter/son, C/D). \textbf{(b)} Post-edit performance metrics: Edit Success (direct change), Logical Generalization (propagation to entailed facts), and Locality (impact on unrelated facts). 
    \textbf{(c)} Correlation between a model's peak bilinear probe accuracy and its maximum logical generalization performance ($R^2=0.939$).
    \textbf{(d)} Layer-wise performance of bilinear probing and logical generalization for ``Not Reversal Cursed'' models.
    }
    \label{fig:edit_generalization}
\vspace{-9pt}
\end{figure}

We evaluated edited models on three metrics (Figure~\ref{fig:edit_generalization}a): (1) \textbf {edit success}, which measures whether the model correctly predicts the primary update (A, husband, B'); (2) \textbf{logical generalization}, defined as the average success rate on the entailed facts: (B', wife, A), (C/D, father, B'), and (B', daughter/son, C/D); and (3) \textbf{locality}, which assesses whether unrelated facts like (C, brother, D) remain unchanged.

\paragraph{Results.} As illustrated in Figure~\ref{fig:edit_generalization}b, both ``Reversal Cursed'' and ``Not Reversal Cursed'' models achieve near-perfect accuracy on the primary edit. However, their capacities for logical propagation diverge sharply: models without the reversal curse exhibit robust generalization to entailed facts, whereas reversal-cursed models fail almost entirely. Furthermore, ``Not Reversal Cursed'' models demonstrate superior locality, preserving the integrity of unrelated factual knowledge more effectively than their counterparts.

To quantify the relationship between internal structure and editing performance, we correlate each model's best bilinear probe accuracy (across all layers from Experiment 2) with its best logical generalization after editing (across all layers). Figure~\ref{fig:edit_generalization}c reveals a robust positive correlation ($R^2=0.939$), establishing that a well-structured bilinear representation predicts a successful logical propagation of a single edit.

Interestingly, the optimal layers for editing do not align perfectly with the layers where the bilinear structure is strongest. Figure~\ref{fig:edit_generalization}d reveals that for ``Not Reversal Cursed'' models, logical generalization is highest when editing early-to-mid layers (1-4), whereas the bilinear structure is most prominent in middle layers (6-9). This suggests that to effectively edit an entity, one must intervene at the layers where the structured representation is being formed, rather than at the layers where it is already fully established and utilized. Modifying these earlier layers appears to correctly update the downstream representation, enabling the desired logical propagation. These results provide strong evidence for our central claim: bilinear representation is a key to logically consistent model editing.

\section{Discussion}

Our findings establish a clear mechanistic link between relational structures in training data, emergent representational geometry, and logical generalization. These results have significant implications for how we understand, build, and control language models.

\paragraph{Model Editing: Is the Model ``Ready'' to be Edited?}
Our results reframe the fundamental challenge of model editing. Much of the current research focuses on developing more sophisticated editing algorithms, treating the model as a static object to be operated upon. We show that the success of any edit is fundamentally constrained by the model's pre-existing representational geometry: an editor cannot force a logically entailed update if the underlying architecture lacks the necessary algebraic structure required to represent that entailment. 

However, the existence of structure alone is not enough; the editing method must also preserve this structural integrity. While we show that fine-tuning can successfully propagate edits when a bilinear structure is present, recent evidence suggests that aggressive editing methods can ``shatter'' the underlying graph topology~\citep{nishi2025representation}. Thus, while bilinear structure is a key indicator of logical generalization, realizing this potential requires editing algorithms that respect—rather than destroy—the model's internal algebraic integrity. This suggests a paradigm shift: before editing, one must first assess whether a model is ``editable'' in a logically consistent way. This leads to a two-pronged approach for future research: 1) the development of geometry-aware editing algorithms that leverage internal relational structures, and 2) ``architectural preparation'' strategies that explicitly endow models with structured knowledge representations during pre-training or fine-tuning.

% \paragraph{The Double-Edged Sword of Structured Knowledge.}
% While a well-structured bilinear representation enables the desirable propagation of edits, it also presents a safety challenge. In a model with this structure, a single edit could trigger a cascade of unintended changes throughout its knowledge base, a phenomenon we might call ``catastrophic generalization.'' For example, consider a model trained on data before 2025. Updating it with a single edit changing ``The current US president is Joe Biden'' to ``The current US president is Donald Trump'' could be disastrous if the model's generalization is too broad. It may incorrectly rewrite dozens of related facts from 2020 to 2024, presenting a distorted and false view of the world to the user. Ensuring safe model editing will therefore require promoting logical generalization as well as developing mechanisms to trace, constrain, and verify the downstream impact of any modification within this tightly coupled algebraic system.

\paragraph{From Memorization to Reasoning: The Role of Representation.}
This work suggests that the perceived gap between a model's ability to memorize and its ability to reason may be a function of its internal knowledge structure. Phenomena like the reversal curse can be viewed as consequences of optimizing models to exploit statistical shortcuts (e.g., term co-occurrence in the training data) rather than to develop capabilities that support logical understanding, such as latent multi-hop reasoning~\citep{yang2024large2,yang2024large,balesni2025lesson}. Our demonstration that a structured knowledge dataset with appropriate regularization can induce a transition to an algebraic, bilinear structure implies that Transformers are capable of learning more than just directional associations. This raises a crucial question for the field: Are we explicitly training models to reason, or are we hoping reasoning emerges as a side effect of scaling? Our results suggest that actively guiding the formation of structured representations, perhaps through curriculum learning, contrastive objectives, or integration with knowledge graphs during pretraining~\citep{yang2025synthetic}, could be a more direct path to building genuinely logical AI.
% \new{\paragraph{Reversal Curse and Model Editing.}
% Most reversal curse studies focus on fine-tuning pretrained models with new facts. \citet{berglund2024the} fine-tuned pretrained LLMs with synthetic knowledge (where both subject and object are synthetic entities not in the pretraining data) and demonstrated that the reversal curse appears in this setting. \citet{balesni2025lesson} compared fine-tuning on facts with synthetic subject + synthetic object versus pretrained subject + synthetic object. They found that synthetic+synthetic fine-tuning generalizes poorly, but pretrained+synthetic shows better generalization (latent multi-hop reasoning) for some relations, though they did not evaluate reversal performance. In contrast, \citet{allen-zhu2025physics} and our work study the reversal curse in a pretraining setting. This distinction is crucial: fine-tuning inherits the representational geometry from pretraining, whereas training from scratch allows us to observe how relational structures emerge. Furthermore, our model editing experiments (Experiment 4) demonstrate that the same bilinear structure that enables reversal generalizations during pretraining is also the key to logically consistent editing. This connects our pretraining findings to the practical challenge of editing pretrained models: editing algorithms may fail not because they are poorly designed, but because the model lacks the necessary geometric structure to represent logical entailments.}

\paragraph{Mechanistic Interpretation: The Attention Head Hypothesis.}
While our analysis relies on readout probes and algebraic validation, we hypothesize that the ``bilinear structure'' we detect corresponds to interaction patterns implemented within the self-attention layers. Because attention computes interactions between pairs of tokens through multiplicative comparisons, it is inherently bilinear (e.g., terms of the form $x^\top A y$). It is therefore plausible that particular attention heads are responsible for encoding these relational structures. Recently,~\citet{elhelo2025inferring} have shown that specific attention heads encode relational lookups and tend to appear in mid to late layers, aligning with our observation that bilinear structure emerges in those layers. This suggests that the absence of such specialized attention heads may be a bottleneck in reversal-cursed models. Future work could test this hypothesis by intervening on attention heads whose behavior aligns with our estimated $M_r$ matrices.

\paragraph{Limitations and Future Work.}
Our primary limitation is the use of language models trained from scratch on a clean, synthetic dataset. This raises the crucial question of whether these findings scale to large, pre-trained language models and the noisy, complex knowledge they contain. Whether similar bilinear structures exist in industrial-scale pre-trained language models remains an open question; it is unlikely that all knowledge is encoded via a single, uniform geometry. Instead, different domains of knowledge may adopt different relational structures in their representations. Our work is a proof of concept, demonstrating that language models are {\it capable} of forming this algebraically robust structure, although we have not verified its prevalence in existing large-scale models. A critical direction for future work, therefore, is to develop methodologies to diagnose {\it how} a pre-trained model decodes a specific piece of information. Such a diagnostic capability would be transformative, enabling a new paradigm of structure-aware model editing. By first identifying a fact's local representational geometry, we could then select or design editing techniques that respect and leverage that structure, moving the field from a trial-and-error process to a more principled, geometrically informed science of knowledge modification.

\section{Conclusion}
\label{sec:conclusion}

We demonstrated that the reversal curse and failures in model editing generalization are not inherent limitations of Transformers, but rather symptoms of an unstructured internal knowledge representation. By training Transformers on a synthetic knowledge graph with appropriate regularization, we showed that they can learn a robust bilinear structure for relational knowledge. Probing experiments confirmed that this structure emerges in the middle layers and is algebraically sound, supporting relational inversion and composition. Crucially, we established a strong link between the presence of this bilinear representation and the model's ability to both overcome the reversal curse and perform logically consistent model editing. When a fact was edited, models with this structure successfully propagated the change to entailed facts, whereas models without it failed to generalize. Our findings highlight that the path toward more reliable and editable language models lies not just in better algorithms but in shaping the fundamental geometry of their learned knowledge representations.

\section*{Reproducibility Statement}
We include complete details about our model architecture, training setup, and hyperparameter sweeps in Appendix~\ref{appendix:training_details}. Our synthetic dataset construction is described in Appendix~\ref{appendix:data}. We ran all experiments on a workstation with 4 A100 GPUs. Our implementation uses the GPT-NeoX library~\citep{gpt-neox-library} via HuggingFace Transformers~\cite{wolf2019huggingface}, which is implemented in PyTorch~\citep{paszke2019pytorch}.

\bibliography{iclr2026_conference}

\begin{thebibliography}{38}
\providecommand{\natexlab}[1]{#1}
\providecommand{\url}[1]{\texttt{#1}}
\expandafter\ifx\csname urlstyle\endcsname\relax
  \providecommand{\doi}[1]{doi: #1}\else
  \providecommand{\doi}{doi: \begingroup \urlstyle{rm}\Url}\fi

\bibitem[Allen-Zhu \& Li(2025)Allen-Zhu and Li]{allen-zhu2025physics}
Zeyuan Allen-Zhu and Yuanzhi Li.
\newblock Physics of language models: Part 3.2, knowledge manipulation.
\newblock In \emph{The Thirteenth International Conference on Learning Representations}, 2025.
\newblock URL \url{https://openreview.net/forum?id=oDbiL9CLoS}.

\bibitem[Andonian et~al.(2023)Andonian, Anthony, Biderman, Black, Gali, Gao, Hallahan, Levy-Kramer, Leahy, Nestler, Parker, Pieler, Phang, Purohit, Schoelkopf, Stander, Songz, Tigges, Thérien, Wang, and Weinbach]{gpt-neox-library}
Alex Andonian, Quentin Anthony, Stella Biderman, Sid Black, Preetham Gali, Leo Gao, Eric Hallahan, Josh Levy-Kramer, Connor Leahy, Lucas Nestler, Kip Parker, Michael Pieler, Jason Phang, Shivanshu Purohit, Hailey Schoelkopf, Dashiell Stander, Tri Songz, Curt Tigges, Benjamin Thérien, Phil Wang, and Samuel Weinbach.
\newblock {GPT-NeoX: Large Scale Autoregressive Language Modeling in PyTorch}, 9 2023.
\newblock URL \url{https://www.github.com/eleutherai/gpt-neox}.

\bibitem[Balesni et~al.(2025)Balesni, Korbak, and Evans]{balesni2025lesson}
Mikita Balesni, Tomek Korbak, and Owain Evans.
\newblock Lessons from studying two-hop latent reasoning, 2025.
\newblock URL \url{https://arxiv.org/abs/2411.16353}.

\bibitem[Berglund et~al.(2024)Berglund, Tong, Kaufmann, Balesni, Stickland, Korbak, and Evans]{berglund2024the}
Lukas Berglund, Meg Tong, Maximilian Kaufmann, Mikita Balesni, Asa~Cooper Stickland, Tomasz Korbak, and Owain Evans.
\newblock The reversal curse: {LLM}s trained on {\textquotedblleft}a is b{\textquotedblright} fail to learn {\textquotedblleft}b is a{\textquotedblright}.
\newblock In \emph{The Twelfth International Conference on Learning Representations}, 2024.
\newblock URL \url{https://openreview.net/forum?id=GPKTIktA0k}.

\bibitem[De~Cao et~al.(2021)De~Cao, Aziz, and Titov]{de-cao2021editing}
Nicola De~Cao, Wilker Aziz, and Ivan Titov.
\newblock Editing factual knowledge in language models.
\newblock In Marie-Francine Moens, Xuanjing Huang, Lucia Specia, and Scott Wen-tau Yih (eds.), \emph{Proceedings of the 2021 Conference on Empirical Methods in Natural Language Processing}, pp.\  6491--6506, Online and Punta Cana, Dominican Republic, November 2021. Association for Computational Linguistics.
\newblock \doi{10.18653/v1/2021.emnlp-main.522}.
\newblock URL \url{https://aclanthology.org/2021.emnlp-main.522/}.

\bibitem[Elhage et~al.(2021)Elhage, Nanda, Olsson, Henighan, Joseph, Mann, Askell, Bai, Chen, Conerly, DasSarma, Drain, Ganguli, Hatfield-Dodds, Hernandez, Jones, Kernion, Lovitt, Ndousse, Amodei, Brown, Clark, Kaplan, McCandlish, and Olah]{elhage2021mathematical}
Nelson Elhage, Neel Nanda, Catherine Olsson, Tom Henighan, Nicholas Joseph, Ben Mann, Amanda Askell, Yuntao Bai, Anna Chen, Tom Conerly, Nova DasSarma, Dawn Drain, Deep Ganguli, Zac Hatfield-Dodds, Danny Hernandez, Andy Jones, Jackson Kernion, Liane Lovitt, Kamal Ndousse, Dario Amodei, Tom Brown, Jack Clark, Jared Kaplan, Sam McCandlish, and Chris Olah.
\newblock A mathematical framework for transformer circuits.
\newblock \emph{Transformer Circuits Thread}, 2021.
\newblock https://transformer-circuits.pub/2021/framework/index.html.

\bibitem[Elhelo \& Geva(2025)Elhelo and Geva]{elhelo2025inferring}
Amit Elhelo and Mor Geva.
\newblock Inferring functionality of attention heads from their parameters.
\newblock In Wanxiang Che, Joyce Nabende, Ekaterina Shutova, and Mohammad~Taher Pilehvar (eds.), \emph{Proceedings of the 63rd Annual Meeting of the Association for Computational Linguistics (Volume 1: Long Papers)}, pp.\  17701--17733, Vienna, Austria, July 2025. Association for Computational Linguistics.
\newblock ISBN 979-8-89176-251-0.
\newblock \doi{10.18653/v1/2025.acl-long.866}.
\newblock URL \url{https://aclanthology.org/2025.acl-long.866/}.

\bibitem[Engels et~al.(2025)Engels, Michaud, Liao, Gurnee, and Tegmark]{engels2025not}
Joshua Engels, Eric~J Michaud, Isaac Liao, Wes Gurnee, and Max Tegmark.
\newblock Not all language model features are one-dimensionally linear.
\newblock In \emph{The Thirteenth International Conference on Learning Representations}, 2025.
\newblock URL \url{https://openreview.net/forum?id=d63a4AM4hb}.

\bibitem[Geva et~al.(2021)Geva, Schuster, Berant, and Levy]{geva2021key}
Mor Geva, Roei Schuster, Jonathan Berant, and Omer Levy.
\newblock Transformer feed-forward layers are key-value memories.
\newblock In Marie-Francine Moens, Xuanjing Huang, Lucia Specia, and Scott Wen-tau Yih (eds.), \emph{Proceedings of the 2021 Conference on Empirical Methods in Natural Language Processing}, pp.\  5484--5495, Online and Punta Cana, Dominican Republic, November 2021. Association for Computational Linguistics.
\newblock \doi{10.18653/v1/2021.emnlp-main.446}.
\newblock URL \url{https://aclanthology.org/2021.emnlp-main.446/}.

\bibitem[Geva et~al.(2023)Geva, Bastings, Filippova, and Globerson]{geva2023dissecting}
Mor Geva, Jasmijn Bastings, Katja Filippova, and Amir Globerson.
\newblock Dissecting recall of factual associations in auto-regressive language models.
\newblock In \emph{Proceedings of the 2023 Conference on Empirical Methods in Natural Language Processing}, pp.\  6109--6125, 2023.

\bibitem[Golovneva et~al.(2024)Golovneva, Allen-Zhu, Weston, and Sukhbaatar]{golovneva2024reverse}
Olga Golovneva, Zeyuan Allen-Zhu, Jason~E Weston, and Sainbayar Sukhbaatar.
\newblock Reverse training to nurse the reversal curse.
\newblock In \emph{First Conference on Language Modeling}, 2024.
\newblock URL \url{https://openreview.net/forum?id=HDkNbfLQgu}.

\bibitem[Gurnee \& Tegmark(2024)Gurnee and Tegmark]{gurnee2024language}
Wes Gurnee and Max Tegmark.
\newblock Language models represent space and time.
\newblock In \emph{The Twelfth International Conference on Learning Representations}, 2024.
\newblock URL \url{https://openreview.net/forum?id=jE8xbmvFin}.

\bibitem[Hase et~al.(2023)Hase, Bansal, Kim, and Ghandeharioun]{hase2023does}
Peter Hase, Mohit Bansal, Been Kim, and Asma Ghandeharioun.
\newblock Does localization inform editing? surprising differences in causality-based localization vs. knowledge editing in language models.
\newblock In \emph{Thirty-seventh Conference on Neural Information Processing Systems}, 2023.
\newblock URL \url{https://openreview.net/forum?id=EldbUlZtbd}.

\bibitem[Hase et~al.(2024)Hase, Hofweber, Zhou, Stengel-Eskin, and Bansal]{hase2024fundamental}
Peter Hase, Thomas Hofweber, Xiang Zhou, Elias Stengel-Eskin, and Mohit Bansal.
\newblock Fundamental problems with model editing: How should rational belief revision work in {LLM}s?
\newblock \emph{Transactions on Machine Learning Research}, 2024.
\newblock ISSN 2835-8856.
\newblock URL \url{https://openreview.net/forum?id=LRf19n5Ly3}.

\bibitem[Hernandez et~al.(2024)Hernandez, Sharma, Haklay, Meng, Wattenberg, Andreas, Belinkov, and Bau]{hernandez2024linearity}
Evan Hernandez, Arnab~Sen Sharma, Tal Haklay, Kevin Meng, Martin Wattenberg, Jacob Andreas, Yonatan Belinkov, and David Bau.
\newblock Linearity of relation decoding in transformer language models.
\newblock In \emph{The Twelfth International Conference on Learning Representations}, 2024.
\newblock URL \url{https://openreview.net/forum?id=w7LU2s14kE}.

\bibitem[Kitouni et~al.(2024)Kitouni, Nolte, Williams, Rabbat, Bouchacourt, and Ibrahim]{kitouni2024the}
Ouail Kitouni, Niklas Nolte, Adina Williams, Michael Rabbat, Diane Bouchacourt, and Mark Ibrahim.
\newblock The factorization curse: Which tokens you predict underlie the reversal curse and more.
\newblock In \emph{The Thirty-eighth Annual Conference on Neural Information Processing Systems}, 2024.
\newblock URL \url{https://openreview.net/forum?id=f70e6YYFHF}.

\bibitem[Lampinen et~al.(2025)Lampinen, Chaudhry, Chan, Wild, Wan, Ku, Bornschein, Pascanu, Shanahan, and McClelland]{lampinen2025generalization}
Andrew~K Lampinen, Arslan Chaudhry, Stephanie~CY Chan, Cody Wild, Diane Wan, Alex Ku, J{\"o}rg Bornschein, Razvan Pascanu, Murray Shanahan, and James~L McClelland.
\newblock On the generalization of language models from in-context learning and finetuning: a controlled study.
\newblock \emph{arXiv preprint arXiv:2505.00661}, 2025.

\bibitem[Lin et~al.(2024)Lin, Fu, Liu, Xie, Lin, Wang, Cai, Wu, and Ye]{lin2024delving}
Zhengkai Lin, Zhihang Fu, Kai Liu, Liang Xie, Binbin Lin, Wenxiao Wang, Deng Cai, Yue Wu, and Jieping Ye.
\newblock Delving into the reversal curse: How far can large language models generalize?
\newblock In \emph{The Thirty-eighth Annual Conference on Neural Information Processing Systems}, 2024.
\newblock URL \url{https://openreview.net/forum?id=1wxFznQWhp}.

\bibitem[Loshchilov \& Hutter(2019)Loshchilov and Hutter]{loshchilov2018decoupled}
Ilya Loshchilov and Frank Hutter.
\newblock Decoupled weight decay regularization.
\newblock In \emph{International Conference on Learning Representations}, 2019.
\newblock URL \url{https://openreview.net/forum?id=Bkg6RiCqY7}.

\bibitem[Meng et~al.(2022)Meng, Bau, Andonian, and Belinkov]{meng2022locating}
Kevin Meng, David Bau, Alex~J Andonian, and Yonatan Belinkov.
\newblock Locating and editing factual associations in {GPT}.
\newblock In \emph{Advances in Neural Information Processing Systems}, 2022.
\newblock URL \url{https://openreview.net/forum?id=-h6WAS6eE4}.

\bibitem[Meng et~al.(2023)Meng, Sharma, Andonian, Belinkov, and Bau]{meng2023massediting}
Kevin Meng, Arnab~Sen Sharma, Alex~J Andonian, Yonatan Belinkov, and David Bau.
\newblock Mass-editing memory in a transformer.
\newblock In \emph{The Eleventh International Conference on Learning Representations}, 2023.
\newblock URL \url{https://openreview.net/forum?id=MkbcAHIYgyS}.

\bibitem[Merullo et~al.(2024)Merullo, Eickhoff, and Pavlick]{merullo2024language}
Jack Merullo, Carsten Eickhoff, and Ellie Pavlick.
\newblock Language models implement simple {W}ord2{V}ec-style vector arithmetic.
\newblock In Kevin Duh, Helena Gomez, and Steven Bethard (eds.), \emph{Proceedings of the 2024 Conference of the North American Chapter of the Association for Computational Linguistics: Human Language Technologies (Volume 1: Long Papers)}, pp.\  5030--5047, Mexico City, Mexico, June 2024. Association for Computational Linguistics.
\newblock \doi{10.18653/v1/2024.naacl-long.281}.
\newblock URL \url{https://aclanthology.org/2024.naacl-long.281/}.

\bibitem[Mikolov et~al.(2013)Mikolov, Sutskever, Chen, Corrado, and Dean]{mikolov2013word2vec}
Tomas Mikolov, Ilya Sutskever, Kai Chen, Greg~S Corrado, and Jeff Dean.
\newblock Distributed representations of words and phrases and their compositionality.
\newblock In C.J. Burges, L.~Bottou, M.~Welling, Z.~Ghahramani, and K.Q. Weinberger (eds.), \emph{Advances in Neural Information Processing Systems}, volume~26. Curran Associates, Inc., 2013.
\newblock URL \url{https://proceedings.neurips.cc/paper_files/paper/2013/file/9aa42b31882ec039965f3c4923ce901b-Paper.pdf}.

\bibitem[Nickel(2013)]{nickel2013tensor}
Maximilian Nickel.
\newblock \emph{Tensor factorization for relational learning}.
\newblock Ludwig-Maximilians-Universit{\"a}t M{\"u}nchen, August 2013.
\newblock URL \url{http://nbn-resolving.de/urn:nbn:de:bvb:19-160568}.

\bibitem[Nickel et~al.(2011)Nickel, Tresp, and Kriegel]{nickel2011three}
Maximilian Nickel, Volker Tresp, and Hans-Peter Kriegel.
\newblock A three-way model for collective learning on multi-relational data.
\newblock In \emph{Proceedings of the 28th international conference on machine learning (ICML-11)}, pp.\  809--816, 2011.

\bibitem[Nishi et~al.(2025)Nishi, Ramesh, Okawa, Khona, Tanaka, and Lubana]{nishi2025representation}
Kento Nishi, Rahul Ramesh, Maya Okawa, Mikail Khona, Hidenori Tanaka, and Ekdeep~Singh Lubana.
\newblock Representation shattering in transformers: A synthetic study with knowledge editing.
\newblock In \emph{Forty-second International Conference on Machine Learning}, 2025.
\newblock URL \url{https://openreview.net/forum?id=BKOeyZal0x}.

\bibitem[Paccanaro \& Hinton(2002)Paccanaro and Hinton]{paccanaro2002learning}
Alberto Paccanaro and Geoffrey~E. Hinton.
\newblock Learning distributed representations of concepts using linear relational embedding.
\newblock \emph{IEEE Transactions on Knowledge and Data Engineering}, 13\penalty0 (2):\penalty0 232--244, 2002.

\bibitem[Pan et~al.(2025)Pan, Wang, Cao, Shi, Yang, Li, and Wang]{pan2025precise}
Haowen Pan, Xiaozhi Wang, Yixin Cao, Zenglin Shi, Xun Yang, Juanzi Li, and Meng Wang.
\newblock Precise localization of memories: A fine-grained neuron-level knowledge editing technique for {LLM}s.
\newblock In \emph{The Thirteenth International Conference on Learning Representations}, 2025.
\newblock URL \url{https://openreview.net/forum?id=5xP1HDvpXI}.

\bibitem[Paszke et~al.(2019)Paszke, Gross, Massa, Lerer, Bradbury, Chanan, Killeen, Lin, Gimelshein, Antiga, Desmaison, Kopf, Yang, DeVito, Raison, Tejani, Chilamkurthy, Steiner, Fang, Bai, and Chintala]{paszke2019pytorch}
Adam Paszke, Sam Gross, Francisco Massa, Adam Lerer, James Bradbury, Gregory Chanan, Trevor Killeen, Zeming Lin, Natalia Gimelshein, Luca Antiga, Alban Desmaison, Andreas Kopf, Edward Yang, Zachary DeVito, Martin Raison, Alykhan Tejani, Sasank Chilamkurthy, Benoit Steiner, Lu~Fang, Junjie Bai, and Soumith Chintala.
\newblock Pytorch: An imperative style, high-performance deep learning library.
\newblock In H.~Wallach, H.~Larochelle, A.~Beygelzimer, F.~d\textquotesingle Alch\'{e}-Buc, E.~Fox, and R.~Garnett (eds.), \emph{Advances in Neural Information Processing Systems}, volume~32. Curran Associates, Inc., 2019.
\newblock URL \url{https://proceedings.neurips.cc/paper_files/paper/2019/file/bdbca288fee7f92f2bfa9f7012727740-Paper.pdf}.

\bibitem[Thibodeau(2022)]{thibodeau2022but}
Jacques Thibodeau.
\newblock But is it really in rome? an investigation of the rome model editing technique.
\newblock \emph{Alignment Forum}, 2022.
\newblock URL \url{https://www.alignmentforum.org/posts/QL7J9wmS6W2fWpofd/but-is-it-really-in-rome-an-investigation-of-the-rome-model}.

\bibitem[Wang et~al.(2024)Wang, Zhang, Tian, Xi, Yao, Xu, Wang, Mao, Wang, Cheng, Liu, Ni, Zheng, and Chen]{wang2024easyedit}
Peng Wang, Ningyu Zhang, Bozhong Tian, Zekun Xi, Yunzhi Yao, Ziwen Xu, Mengru Wang, Shengyu Mao, Xiaohan Wang, Siyuan Cheng, Kangwei Liu, Yuansheng Ni, Guozhou Zheng, and Huajun Chen.
\newblock {E}asy{E}dit: An easy-to-use knowledge editing framework for large language models.
\newblock In Yixin Cao, Yang Feng, and Deyi Xiong (eds.), \emph{Proceedings of the 62nd Annual Meeting of the Association for Computational Linguistics (Volume 3: System Demonstrations)}, pp.\  82--93, Bangkok, Thailand, August 2024. Association for Computational Linguistics.
\newblock \doi{10.18653/v1/2024.acl-demos.9}.
\newblock URL \url{https://aclanthology.org/2024.acl-demos.9/}.

\bibitem[Wolf et~al.(2019)Wolf, Debut, Sanh, Chaumond, Delangue, Moi, Cistac, Rault, Louf, Funtowicz, et~al.]{wolf2019huggingface}
Thomas Wolf, Lysandre Debut, Victor Sanh, Julien Chaumond, Clement Delangue, Anthony Moi, Pierric Cistac, Tim Rault, R{\'e}mi Louf, Morgan Funtowicz, et~al.
\newblock Huggingface's transformers: State-of-the-art natural language processing.
\newblock \emph{arXiv preprint arXiv:1910.03771}, 2019.

\bibitem[Yang et~al.(2024)Yang, Gribovskaya, Kassner, Geva, and Riedel]{yang2024large2}
Sohee Yang, Elena Gribovskaya, Nora Kassner, Mor Geva, and Sebastian Riedel.
\newblock Do large language models latently perform multi-hop reasoning?
\newblock In Lun-Wei Ku, Andre Martins, and Vivek Srikumar (eds.), \emph{Proceedings of the 62nd Annual Meeting of the Association for Computational Linguistics (Volume 1: Long Papers)}, pp.\  10210--10229, Bangkok, Thailand, August 2024. Association for Computational Linguistics.
\newblock \doi{10.18653/v1/2024.acl-long.550}.
\newblock URL \url{https://aclanthology.org/2024.acl-long.550/}.

\bibitem[Yang et~al.(2025{\natexlab{a}})Yang, Kassner, Gribovskaya, Riedel, and Geva]{yang2024large}
Sohee Yang, Nora Kassner, Elena Gribovskaya, Sebastian Riedel, and Mor Geva.
\newblock Do large language models perform latent multi-hop reasoning without exploiting shortcuts?
\newblock In Wanxiang Che, Joyce Nabende, Ekaterina Shutova, and Mohammad~Taher Pilehvar (eds.), \emph{Findings of the Association for Computational Linguistics: ACL 2025}, Vienna, Austria, July 2025{\natexlab{a}}. Association for Computational Linguistics.
\newblock \doi{10.18653/v1/2025.findings-acl.205}.
\newblock URL \url{https://aclanthology.org/2025.findings-acl.205/}.

\bibitem[Yang et~al.(2025{\natexlab{b}})Yang, Band, Li, Candes, and Hashimoto]{yang2025synthetic}
Zitong Yang, Neil Band, Shuangping Li, Emmanuel Candes, and Tatsunori Hashimoto.
\newblock Synthetic continued pretraining.
\newblock In \emph{The Thirteenth International Conference on Learning Representations}, 2025{\natexlab{b}}.
\newblock URL \url{https://openreview.net/forum?id=07yvxWDSla}.

\bibitem[Yao et~al.(2023)Yao, Wang, Tian, Cheng, Li, Deng, Chen, and Zhang]{yao2023editing}
Yunzhi Yao, Peng Wang, Bozhong Tian, Siyuan Cheng, Zhoubo Li, Shumin Deng, Huajun Chen, and Ningyu Zhang.
\newblock Editing large language models: Problems, methods, and opportunities.
\newblock In Houda Bouamor, Juan Pino, and Kalika Bali (eds.), \emph{Proceedings of the 2023 Conference on Empirical Methods in Natural Language Processing}, pp.\  10222--10240, Singapore, December 2023. Association for Computational Linguistics.
\newblock \doi{10.18653/v1/2023.emnlp-main.632}.
\newblock URL \url{https://aclanthology.org/2023.emnlp-main.632/}.

\bibitem[Zhu et~al.(2020)Zhu, Rawat, Zaheer, Bhojanapalli, Li, Yu, and Kumar]{zhu2020modifying}
Chen Zhu, Ankit~Singh Rawat, Manzil Zaheer, Srinadh Bhojanapalli, Daliang Li, Felix Yu, and Sanjiv Kumar.
\newblock Modifying memories in transformer models.
\newblock \emph{arXiv preprint arXiv:2012.00363}, 2020.

\bibitem[Zhu et~al.(2024)Zhu, Huang, Zhang, Jordan, Jiao, Tian, and Russell]{zhu2024towards}
Hanlin Zhu, Baihe Huang, Shaolun Zhang, Michael Jordan, Jiantao Jiao, Yuandong Tian, and Stuart Russell.
\newblock Towards a theoretical understanding of the 'reversal curse' via training dynamics.
\newblock In \emph{The Thirty-eighth Annual Conference on Neural Information Processing Systems}, 2024.
\newblock URL \url{https://openreview.net/forum?id=QoWf3lo6m7}.

\end{thebibliography}
\bibliographystyle{iclr2026_conference}

\newpage
\appendix

\section{Model Architecture and Training Details}
\label{appendix:training_details}
We use a decoder-only Transformer (GPT-NeoX) with rotary positional embeddings (RoPE).

Architecture configuration (GPT-NeoX):
\begin{itemize}
    \item Layers: 12 Transformer blocks
    \item Hidden size (or embedding dimension): 896
    \item Attention heads: 16 (head dimension 56)
    \item Feed-forward size: 3584 (4$\times$ hidden size)
    \item Positional encoding: rotary embeddings with standard base $10{,}000$
    \item Max context length: 1024 tokens
    \item Dropout: attention 0.1, MLP hidden 0.1
    \item Residual path: non-parallel residual (\texttt{use\_parallel\_residual = False})
    \item Number of parameters: $\sim$206M.
\end{itemize}

Training setup:
\begin{itemize}
    \item Hardware: 4 A100 GPUs
    \item Batch size: per-device 16 (train), 32 (eval); global 64 (train), 128 (eval)
    \item Optimizer: AdamW with learning rate $3\times 10^{-4}$ and $(\beta_1,\beta_2)=(0.9, 0.95)$.
    \item Hyperparameter sweep: weight decay $\in \{0, 0.1, 0.5, 1, 2, 3, 4, 5, 6\}$; random seed $\in \{0,1,2\}$
    \item Learning rate scheduler: Cosine decay with linear warmup ratio $0.01$
    \item Training epochs: 20
\end{itemize}

Note that we train all models from scratch without using any pretrained weights and we used the tokenizer from GPT-NeoX.

\section{Synthetic Data Construction and Examples}
\label{appendix:data}
We construct a synthetic family-graph dataset where each family contributes a single document formed by concatenating all relational facts as sentences:
“[Subject First Name] [Family Name] [Relation] [Object First Name] [Family Name]”.
A family name is shared by all entities (or members) and is formed by “[Middle Name] [Last Name]”,
so the full name is “[First Name] [Middle Name] [Last Name]”.
We sample names from fixed pools (listed below) to ensure uniqueness and reproducibility.

Generation rules:

\begin{itemize}
    \item Entities: one family per document with unique members; all members share the family name.
    \item Relations: eight types — \texttt{husband}, \texttt{wife}, \texttt{father}, \texttt{mother}, \texttt{brother}, \texttt{sister}, \texttt{son}, \texttt{daughter}.
    \item Split: 1{,}000 families divided into two groups of 500 each.
    \item Training data:
        \begin{itemize}
            \item Group 1 (first 500 families): includes all eight relations. 5{,}000 members total; 36 facts per family. See example below.
            \item Group 2 (next 500 families): excludes \texttt{father}/\texttt{mother}. 5{,}000 members total; 24 facts per family. See example below.
        \end{itemize}
    \item Test data:
        \begin{itemize}
            \item For Group 2 families, add back only the \texttt{father}/\texttt{mother} facts to create the test set (12 facts per family). See example below.
        \end{itemize}
\end{itemize}

\paragraph{Trainset example from the first group (all relations).}
\begingroup\small\ttfamily\begin{flushleft}
Sandy Francis Barton brother Zachary Francis Barton. Katrina Francis Barton son Zachary Francis Barton. Sandy Francis Barton father Kyle Francis Barton. Debra Francis Barton daughter Katrina Francis Barton. Kyle Francis Barton mother Veronica Francis Barton. Kyle Francis Barton daughter Sandy Francis Barton. Debra Francis Barton husband Gary Francis Barton. Henry Francis Barton sister Katrina Francis Barton. Justin Francis Barton wife Veronica Francis Barton. Katrina Francis Barton daughter Sandy Francis Barton. Veronica Francis Barton son Kyle Francis Barton. Vanessa Francis Barton father Justin Francis Barton. Gary Francis Barton son Henry Francis Barton. Gary Francis Barton wife Debra Francis Barton. Kyle Francis Barton father Justin Francis Barton. Gary Francis Barton daughter Katrina Francis Barton. Katrina Francis Barton father Gary Francis Barton. Zachary Francis Barton sister Sandy Francis Barton. Debra Francis Barton son Henry Francis Barton. Zachary Francis Barton father Kyle Francis Barton. Veronica Francis Barton daughter Vanessa Francis Barton. Henry Francis Barton father Gary Francis Barton. Kyle Francis Barton sister Vanessa Francis Barton. Henry Francis Barton mother Debra Francis Barton. Katrina Francis Barton brother Henry Francis Barton. Sandy Francis Barton mother Katrina Francis Barton. Zachary Francis Barton mother Katrina Francis Barton. Vanessa Francis Barton mother Veronica Francis Barton. Katrina Francis Barton husband Kyle Francis Barton. Kyle Francis Barton wife Katrina Francis Barton. Justin Francis Barton son Kyle Francis Barton. Justin Francis Barton daughter Vanessa Francis Barton. Katrina Francis Barton mother Debra Francis Barton. Veronica Francis Barton husband Justin Francis Barton. Vanessa Francis Barton brother Kyle Francis Barton. Kyle Francis Barton son Zachary Francis Barton.
\end{flushleft}\endgroup

\paragraph{Trainset example from the second group (without father/mother).}
\begingroup\small\ttfamily\begin{flushleft}
Dalton Scott Wall sister Colleen Scott Wall. Ebony Scott Wall husband Cody Scott Wall. Ebony Scott Wall son Julian Scott Wall. Jamie Scott Wall brother Julian Scott Wall. Jacob Scott Wall son Dalton Scott Wall. Jacob Scott Wall wife Jamie Scott Wall. Curtis Scott Wall daughter Brenda Scott Wall. Brenda Scott Wall brother Jacob Scott Wall. Emily Scott Wall husband Curtis Scott Wall. Jamie Scott Wall son Dalton Scott Wall. Curtis Scott Wall wife Emily Scott Wall. Cody Scott Wall daughter Jamie Scott Wall. Jamie Scott Wall husband Jacob Scott Wall. Jacob Scott Wall sister Brenda Scott Wall. Emily Scott Wall daughter Brenda Scott Wall. Cody Scott Wall son Julian Scott Wall. Ebony Scott Wall daughter Jamie Scott Wall. Curtis Scott Wall son Jacob Scott Wall. Cody Scott Wall wife Ebony Scott Wall. Colleen Scott Wall brother Dalton Scott Wall. Jamie Scott Wall daughter Colleen Scott Wall. Julian Scott Wall sister Jamie Scott Wall. Jacob Scott Wall daughter Colleen Scott Wall. Emily Scott Wall son Jacob Scott Wall.
\end{flushleft}\endgroup

\paragraph{Testset example from the second group (only father/mother). 12 prompts per family.}
\begingroup\small\ttfamily\begin{flushleft}
Julian Scott Wall mother Ebony Scott Wall. \\
Julian Scott Wall father Cody Scott Wall. \\
Jamie Scott Wall mother Ebony Scott Wall. \\
Jamie Scott Wall father Cody Scott Wall. \\
Jacob Scott Wall mother Emily Scott Wall. \\
Jacob Scott Wall father Curtis Scott Wall. \\
Brenda Scott Wall mother Emily Scott Wall. \\
Brenda Scott Wall father Curtis Scott Wall. \\
Dalton Scott Wall mother Jamie Scott Wall. \\
Dalton Scott Wall father Jacob Scott Wall. \\
Colleen Scott Wall mother Jamie Scott Wall. \\
Colleen Scott Wall father Jacob Scott Wall.\\
\end{flushleft}\endgroup

\paragraph{Name sampling.}
We draw first names by gender, middle names from fixed pools, and last names from a large pool. The family name is “[Middle Name] [Last Name]”, shared by all members. Below are the exact pools used.

\subsection*{Name Pools (for Reproducibility)}
\textbf{FEMALE\_FIRST\_NAMES}
\begingroup\footnotesize\ttfamily\begin{flushleft}
Sheryl, Caitlyn, Alisha, Heidi, Frances, Elaine, Catherine, Bridget, Tami, Norma, Bianca, Robyn, Kylie, Amanda, Alyssa, Brandy, Dorothy, Erica, Melody, Sandra, Alison, Peggy, Debra, Sophia, Victoria, Kristy, Ebony, Loretta, Robin, Holly, Adrienne, Christina, Veronica, Joy, Tasha, Chloe, Doris, Jody, Wanda, Tricia, Kayla, Brenda, Karen, Judith, Sandy, Hailey, Angela, Madeline, Natalie, Carol, Katrina, Beth, Pam, Jamie, Shelia, Sharon, Karina, Rebekah, Deanna, Autumn, Angelica, Ellen, Jade, Sierra, Tracie, Brianna, Susan, Virginia, Lydia, Karla, Christy, Kathleen, Kaitlyn, Diane, Haley, Bailey, Colleen, Nancy, Yesenia, Sara, Madison, Shannon, Hayley, Patty, Terri, Joan, Anne, Emily, Vanessa, Jenny, Kimberly, Hannah, Ashley, Dominique, Rachael, Toni, Melanie, Kerry, Mackenzie, Charlene
\end{flushleft}\endgroup

\textbf{MALE\_FIRST\_NAMES}
\begingroup\footnotesize\ttfamily\begin{flushleft}
Guy, Damon, Gerald, Steve, Samuel, Gregory, Todd, Mark, Timothy, Leroy, Julian, Fernando, Dalton, Rick, Ralph, Cesar, Bill, Clinton, Darren, Dave, Marco, Brandon, Kyle, Kristopher, Noah, Ross, Glen, Shawn, Alec, Cole, Ryan, Harold, Johnathan, Cody, Jacob, Mason, Daryl, Mike, Adam, Wesley, Raymond, Don, Richard, Clayton, Jake, Seth, Edgar, Tracy, Kent, David, Roy, Aaron, Jerome, Phillip, Alexis, Steven, Victor, Javier, Gavin, Brad, Gene, Caleb, Carl, Peter, Brett, Cory, Craig, Jesus, Gary, Oscar, Henry, Cameron, Curtis, Zachary, Mathew, Jared, Ernest, Sergio, Nicholas, Hayden, Kevin, Justin, Jon, Christian, Joseph, Darryl, Eduardo, Joe, Jerry, Duane, Vernon, Micheal, Greg, Frank, Bradley, Corey, Rodney, Angel, Derrick, Terrence
\end{flushleft}\endgroup

\textbf{MIDDLE\_NAMES}
\begingroup\footnotesize\ttfamily\begin{flushleft}
Anthony, Marcus, Jose, Kenneth, Lee, Colin, Arthur, Kirk, Blake, Dan, Benjamin, Marvin, Troy, Philip, Donald, Jamie, Calvin, Luke, Dustin, Marc, Tristan, Andres, Michael, Tyrone, Jeffery, Patrick, Wyatt, Luis, Larry, Frederick, Earl, Darrell, Perry, Roberto, Shannon, Douglas, Eddie, Jaime, Chad, Scott, Norman, Francis, Johnny, Ruben, Bernard, Albert, Rickey, Miguel, Spencer, Brent, Reginald, Leonard, Dennis, Kerry, Ronald, Russell, Gregg, Trevor, Drew, Hunter, Erik, Warren, Jesse, Levi, Francisco, Maxwell, Wayne, Ray, Lonnie, Ricky, Brian, Charles, Parker, Bryce, Bruce, Matthew, Clifford, Edwin, Nathan, Dean, Gordon, Sean, Stanley, Stephen, Karl, Dwayne, Antonio, Brady, Jeffrey, Elijah, Andrew, Adrian, Gilbert, Omar, Taylor, Tanner, Nathaniel, Devin, Lance, Harry
\end{flushleft}\endgroup

\textbf{LAST\_NAMES}
\begingroup\footnotesize\ttfamily\begin{flushleft}
Allison, Hanna, Stark, Mata, Travis, Peters, Zuniga, Smith, Gay, Thornton, Yu, Miller, Webb, Patterson, Ortiz, Combs, Meadows, Christensen, Freeman, Howell, Berger, Cooley, Glover, Jennings, Blackwell, Turner, Mcgee, Duffy, Montgomery, Glenn, Krause, Coleman, Petersen, Gregory, Barnes, Morris, Hensley, Harding, Bird, Estrada, Garza, Gomez, Burke, Waters, Lam, Davenport, Frost, Stafford, Jarvis, Williams,
\end{flushleft}\endgroup

\paragraph{Dataset augmentation.} For each family document, we create augmented training instances by randomly permuting the order of sentences (facts) while keeping each sentence unchanged. We generate 1{,}000 permutations per family, resulting in 318M tokens per training epoch.

\section{SVD-based Update of Relation Matrices in RESCAL}
\label{appendix:rescal_svd}
In the RESCAL model~\citep{nickel2011three}, the Alternating Least Squares (ALS) procedure requires updating the relation matrices $\{\mathbf{M}_r\}_{r=1}^m$ while holding the entity embedding matrix $\mathbf{A}\in\mathbb{R}^{n\times d}$ fixed, where $n$ is the number of entities, $d$ is the embedding dimension, and $m$ is the number of relations.

The objective function for a single relation $r$ is:
\begin{equation}
    L(\mathbf{M}_r) = \frac{1}{2} \| \mathcal{X}_r - \mathbf{A} \mathbf{M}_r \mathbf{A}^T \|_F^2 + \frac{\lambda_R}{2} \|\mathbf{M}_r\|_F^2
    \label{eq:obj_Mr}
\end{equation}
Our goal is to find the $\mathbf{M}_r \in \mathbb{R}^{d \times d}$ that minimizes this function.

To find the minimum, we can set the gradient of $L(\mathbf{M}_r)$ with respect to $\mathbf{M}_r$ to zero:
\begin{equation}
    \frac{\partial L}{\partial \mathbf{M}_r} = -\mathbf{A}^T (\mathcal{X}_r - \mathbf{A} \mathbf{M}_r \mathbf{A}^T) \mathbf{A} + \lambda_R \mathbf{M}_r = 0
\end{equation}
Rearranging the terms, we get the normal equation:
\begin{equation}
    \mathbf{A}^T \mathbf{A} \mathbf{M}_r \mathbf{A}^T \mathbf{A} + \lambda_R \mathbf{M}_r = \mathbf{A}^T \mathcal{X}_r \mathbf{A}
    \label{eq:sylvester}
\end{equation}
This is a continuous Sylvester equation. Using the vectorization operator vec() and the Kronecker product $\otimes$, we can rewrite it as a standard linear system:
\begin{equation}
    \left( (\mathbf{A}^T\mathbf{A}) \otimes (\mathbf{A}^T\mathbf{A}) + \lambda_R \mathbf{I}_{d^2} \right) \text{vec}(\mathbf{M}_r) = \text{vec}(\mathbf{A}^T \mathcal{X}_r \mathbf{A})
    \label{eq:kronecker_form}
\end{equation}
Solving this equation directly requires inverting a dense $(d^2 \times d^2)$ matrix, which is computationally expensive with a complexity of $\mathcal{O}((d^2)^3) = \mathcal{O}(d^6)$. This becomes prohibitive as the embedding dimension $d$ grows. Due to high dimension of embedding space ($d = 896$) of our models, Eq.~\ref{eq:kronecker_form} is infeasible to solve directly.

To mitigate this issue, we employ the Singular Value Decomposition (SVD) to overcome the aforementioned computational bottleneck. Let the SVD of the entity matrix be $\mathbf{A}= \mathbf{U}\mathbf{S}\mathbf{V}^\top$ with
orthonormal $\mathbf{U},\mathbf{V}$ and singular values
$\mathbf{S}=\operatorname{diag}(s_1,\dots,s_d)$.
Using $\mathbf{A}^\top\mathbf{A}= \mathbf{V}\mathbf{S}^2\mathbf{V}^\top$,
Eq.~\ref{eq:sylvester} can be rotated into the singular space,
yielding the diagonal Sylvester equation
\[
  s_i^2(\tilde{\mathbf{M}}_r)_{ij}s_j^2+\lambda_R(\tilde{\mathbf{M}}_r)_{ij}
  =s_i(\tilde{\mathcal{X}}_r)_{ij}s_j ,
\qquad
\tilde{\mathbf{M}}_r=\mathbf{V}^\top\mathbf{M}_r\mathbf{V},\;
\tilde{\mathcal{X}}_r=\mathbf{U}^\top\mathcal{X}_r\mathbf{U}.
\]
Solving element-wise gives
\[
  (\tilde{\mathbf{M}}_r)_{ij}= \frac{s_i s_j}{s_i^{2}s_j^{2}+\lambda_R}\,
                               (\tilde{\mathcal{X}}_r)_{ij},
\]
or in matrix form $\tilde{\mathbf{M}}_r=\mathbf{P}\odot\tilde{\mathcal{X}}_r$
with $P_{ij}=s_i s_j /(s_i^{2}s_j^{2}+\lambda_R)$.
Transforming back,
\[
  \boxed{\;
  \mathbf{M}_r
  =\mathbf{V}\bigl(\mathbf{P}\odot(\mathbf{U}^\top\mathcal{X}_r\mathbf{U})\bigr)\mathbf{V}^\top }.
\]

The update costs $\mathcal{O}(nd^{2})$ for the SVD and
$\mathcal{O}(n^{2}d+d^{3})$ for the remaining multiplications,
vs.\ $\mathcal{O}(d^{6})$ for the naive Kronecker inversion.
In this work, we employed the algorithm in \citet{nickel2013tensor} and reproduce the formula here only to motivate our implementation choice.

$n=1250$ (max entities; 125 families, 10 members per family) and $d=896$ (embedding dimension) in our experiments, making the SVD-based update feasible. $M_r$ is obtained from train set $A$ then evaluated on test set $B$. $B$ and $A$ have disjoint entity sets.

\newpage
\section{Training, probing, and editing results}
\label{appendix:results}

\subsection{Training result in detail}
\label{appendix:train_results}
\begin{figure}[H]
    \centering
    \includegraphics[width=1.0\textwidth]{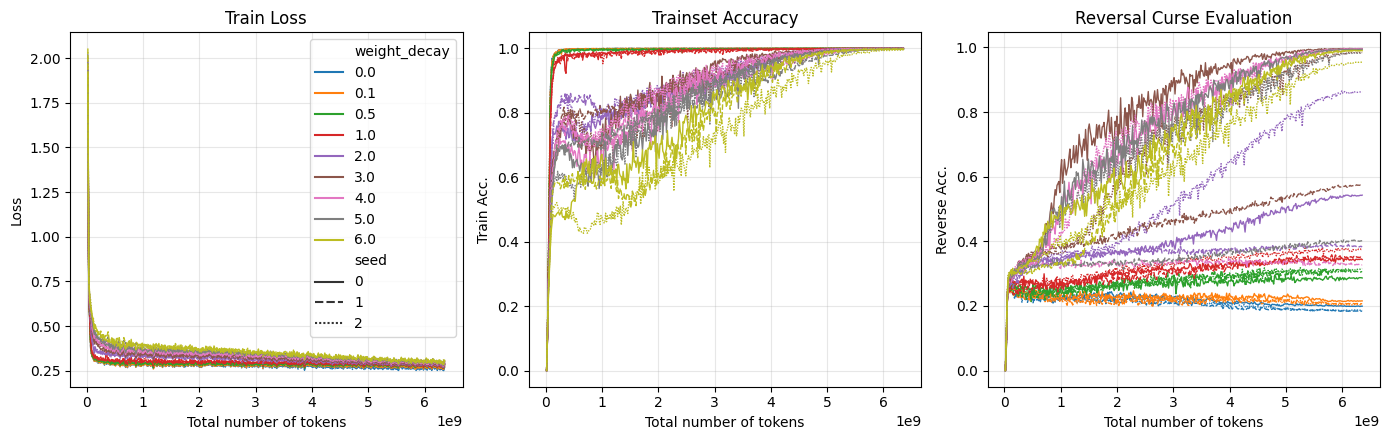}
    \caption{Training loss, train accuracy, and test accuracy for models with different weight decay values and seeds.}
    \label{sfig:train_loss}
\end{figure}

In this section, we provide training curves for all models with different weight decay values and random seeds over training steps (see Fig~\ref{sfig:train_loss}). All models achieve 100\% training accuracy, but test accuracy varies significantly based on weight decay and random seed.

\subsection{Probing results in detail}

\subsubsection{Linear relation embedding}
\label{appendix:lre_N}
\begin{figure}[!b]
    \centering
    \includegraphics[width=0.95\textwidth]{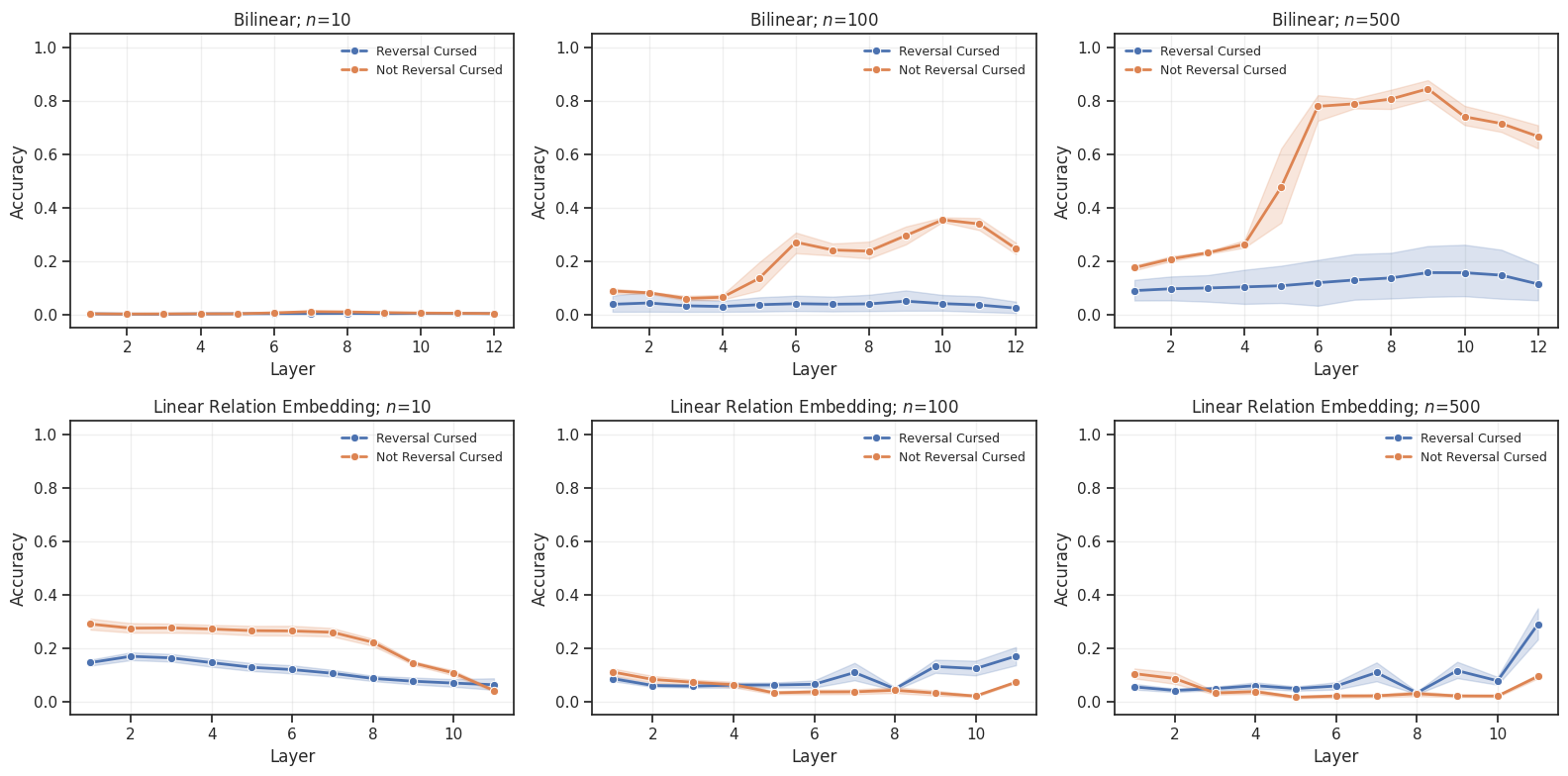}
    \caption{Bilinear and linear relational embedding accuracy for $n=10$, $n=100$, and $n=500$, where $n$ denotes the number of training samples per relation.}
    \label{fig:lre_n10_100}
\end{figure}

Figure \ref{fig:lre_n10_100} shows the probe’s accuracy as a function of the number of training samples per relation, $n$ ($n=10$, $n=100$, and $n=500$). The main text reports results for $n=10$ due to the high computational cost of Jacobian calculations. Here, we show that increasing the number of samples up to $n=500$ does not improve performance, confirming that the poor accuracy of the linear relational embedding probe is not due to insufficient sampling. We also run our bilinear probing for $n=10$, $n=100$, and $n=500$, which shows that insufficient $n$ leads to underfitting.

\begin{figure}[!t]
    \centering
    \includegraphics[width=1.0\textwidth]{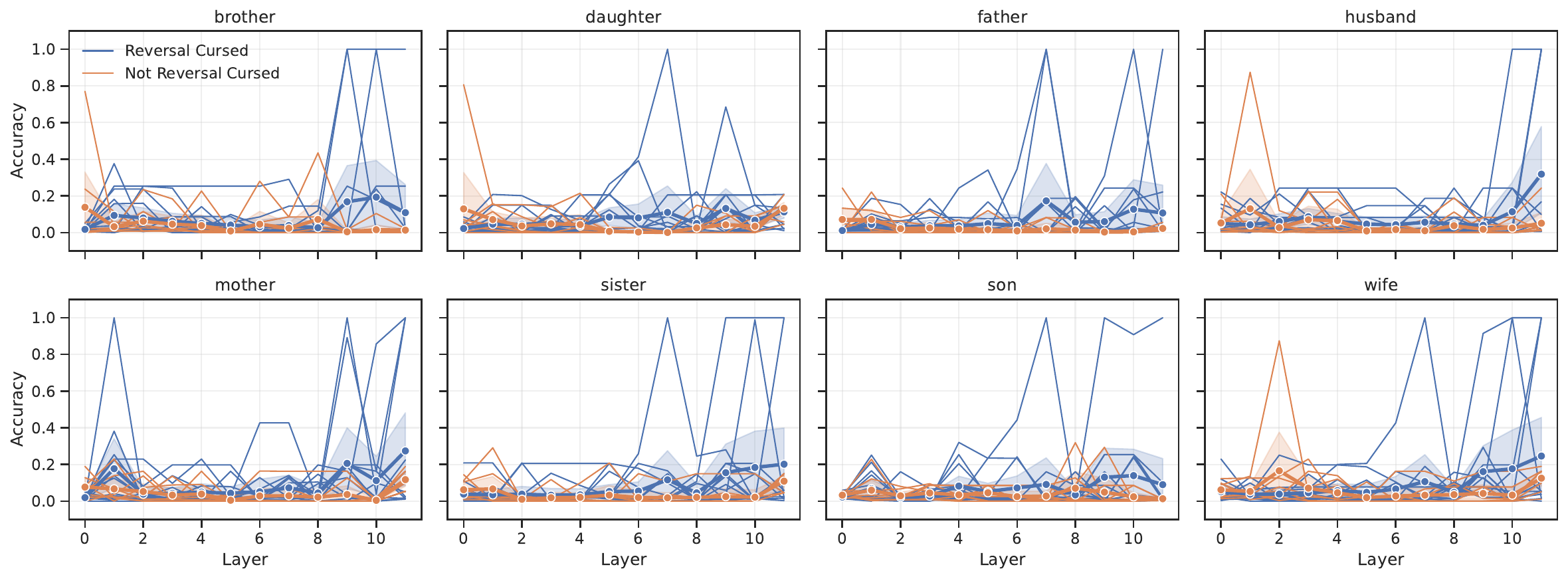}
    \caption{Visualization of linear relational embedding probing results for each relation $r$ as a spaghetti plot ($n=100$ and $\beta=5$).}
    \label{fig:lre_spaghetti}
\end{figure}

Figure~\ref{fig:lre_spaghetti} visualizes the layer-wise accuracy for each of the eight relations individually for $N=100$ and $\beta=5$. Interestingly, few models in ``Reversal cursed'' group (blue) show high accuracy at mid-late layers while ``Not Reversal Cursed'' models (orange) do not. It indicates that some models in the ``Reversal cursed'' group do learn linear relational embeddings for certain relations, but they are not consistent across relations and layers.

\subsubsection{Translational}

\begin{figure}[!t]
    \centering
    \includegraphics[width=1.0\textwidth]{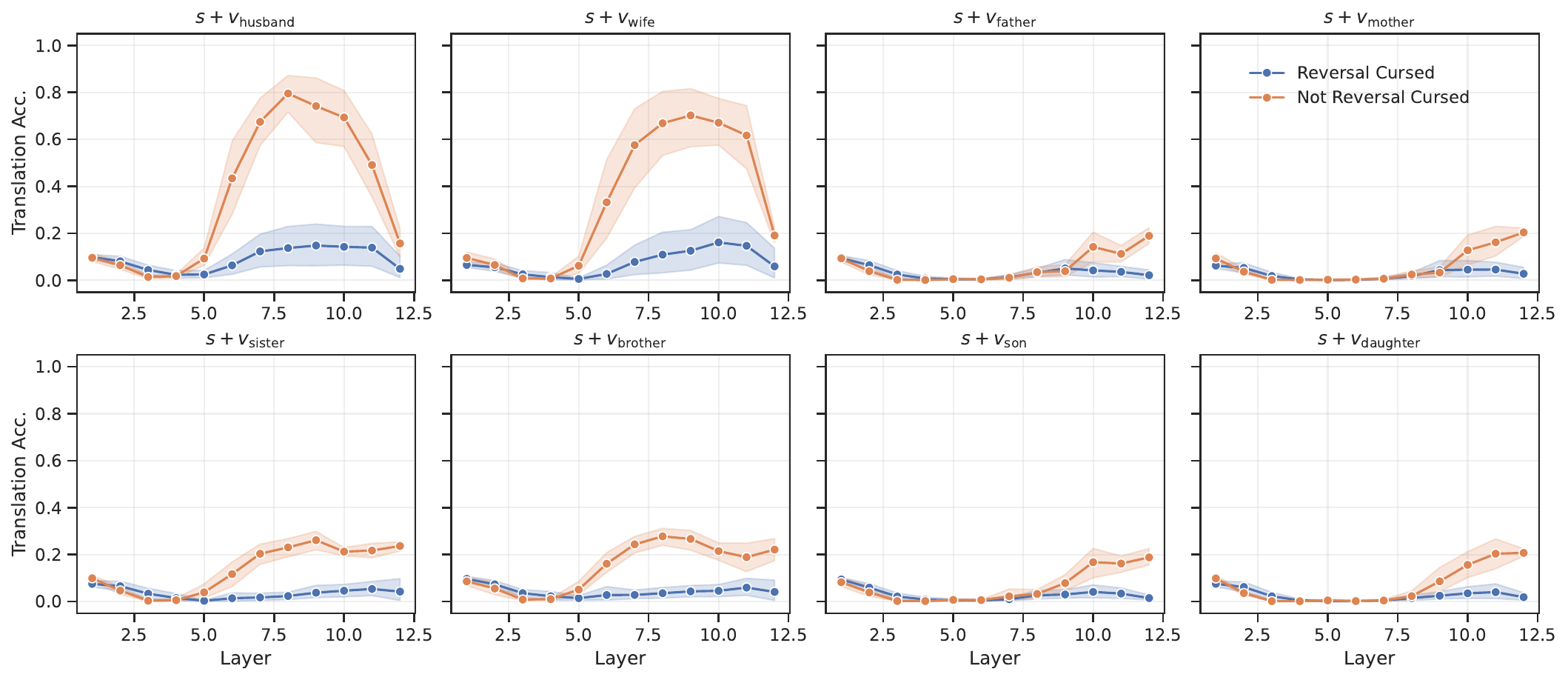}
    \caption{Translational probing accuracy for each relation $r$.}
    \label{fig:translation_relations}
\end{figure}

Figure~\ref{fig:translation_relations} shows the per-relation accuracy for this task. The ``Not Reversal Cursed'' models exhibit high accuracy only for the symmetric \texttt{husband} and \texttt{wife} relations, peaking at mid-to-late layers. Accuracy for all other relations is near zero for both model groups. This suggests that while a translational structure is learned, it is limited to simple symmetric pairs and does not generalize to other relation types.

% We further test if the learned relation vectors capture inverse properties, i.e., $v_r \approx -v_{r^{-1}}$. Figure~\ref{fig:translation_negation} shows the accuracy for predicting an object by subtracting the inverse relation vector (e.g., predicting a husband via $ - v_{\text{wife}}$). The ``Not Reversal Cursed'' models again show high accuracy for the \texttt{husband}/\texttt{wife} pair, indicating they learn that $v_{\text{husband}} \approx -v_{\text{wife}}$. This property is weaker for the \texttt{brother}/\texttt{sister} pair and absent in ``Reversal Cursed'' models, reinforcing that the learned translational geometry is sparse and strongly correlated with overcoming the reversal curse.

% \begin{figure}
%     \centering
%     \includegraphics[width=1.0\textwidth]{SFigs/SM-translation_negation.pdf}
%     \caption{Translation accuracy under negation.}
%     \label{fig:translation_negation}
% \end{figure}

\subsubsection{Bilinear}

\begin{figure}
    \centering
    \includegraphics[width=1.0\textwidth]{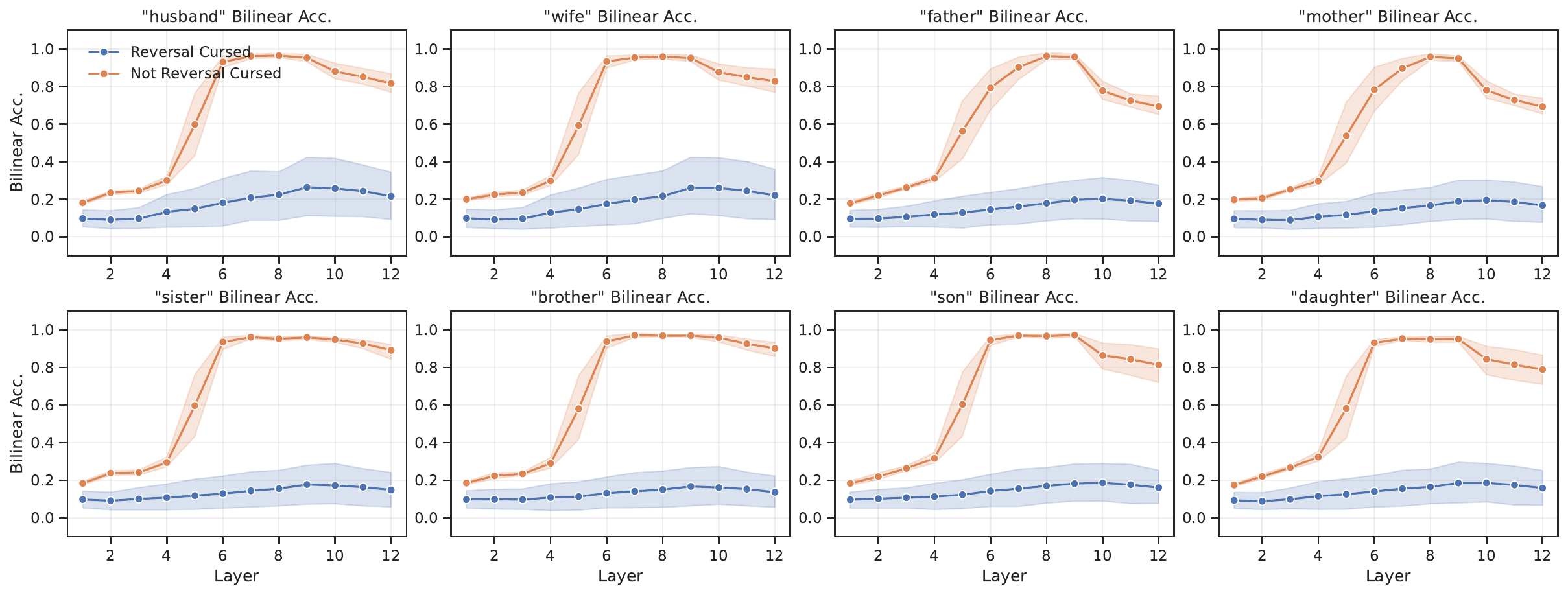}
    \caption{Bilinear probing accuracy for each relation $r$.}
    \label{fig:rescal_relations}
\end{figure}

Figure~\ref{fig:rescal_relations} shows the per-relation accuracy of bilinear probing. The ``Not Reversal Cursed'' models achieve high accuracy across all relations, peaking at mid-to-late layers. In contrast, ``Reversal Cursed'' models show near-zero accuracy for all relations. This indicates that learning a bilinear relational structure is strongly associated with overcoming the reversal curse and generalizes well across different relation types.

\subsection{Editing result in detail}
\label{appendix:editing}

\textbf{Setup.} For each model we sample 50 distinct husband facts (A, husband, B)  from Group~1. Each is edited to (A, husband, B') where B' is another female entity from the \emph{same} family (preserves name template and type). Single edit per run; no simultaneous multi-fact changes. For every layer $l\in\{1,\dots,12\}$ we fine-tune only the MLP output (final linear) weights of that layer using a single example $(A,\texttt{husband},B')$ . Optimizer: Adam, lr $4\times10^{-4}$, early stop when loss $<0.2$ (cap 50 update steps). All other parameters frozen.

\begin{figure}
    \centering
    \includegraphics[width=1.0\textwidth]{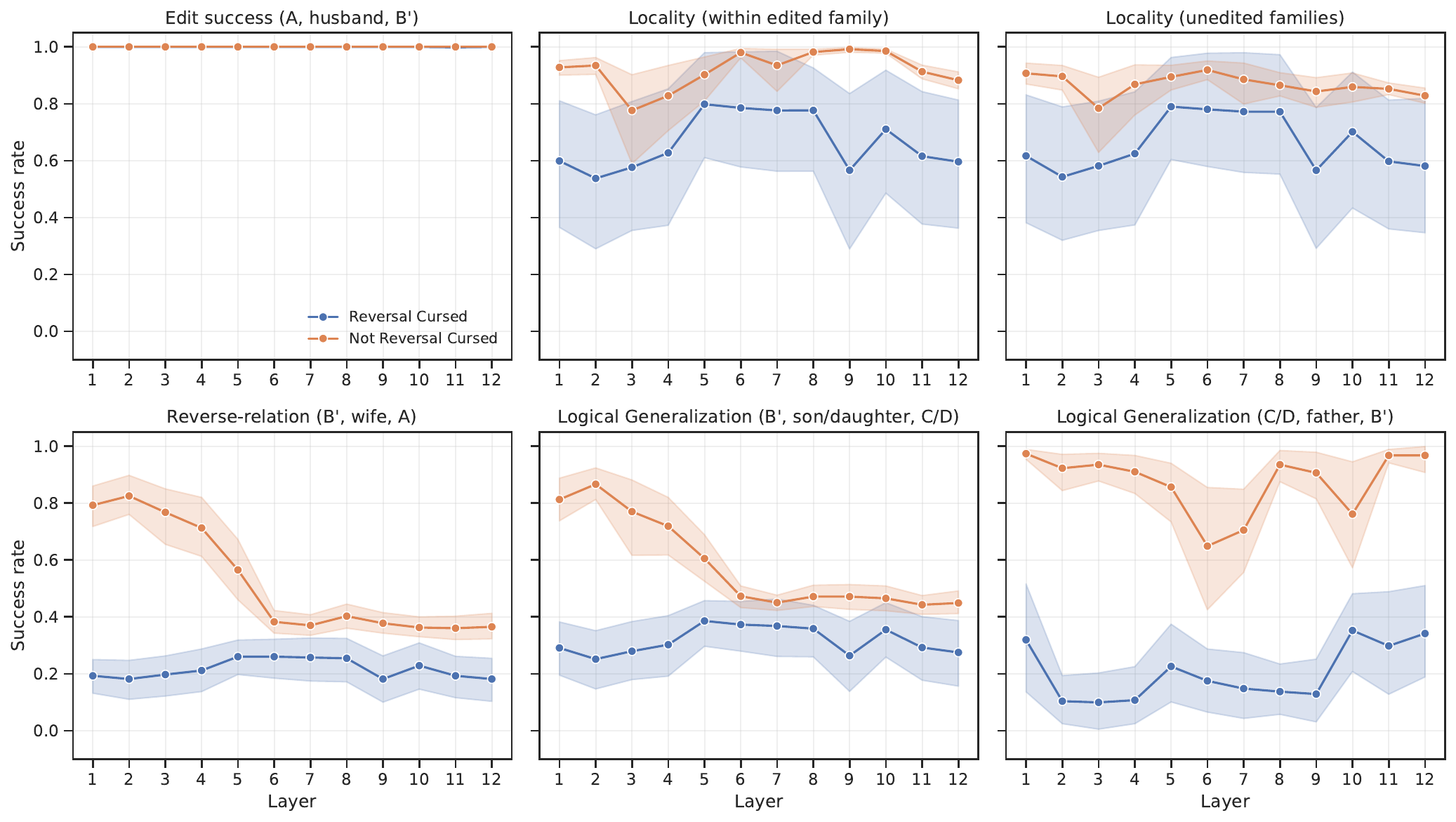}
    \caption{Editing experiment details. Six panels: Edit Success, Locality (edited family), Locality (other families), Reverse relation, Logical Generalization to children, Logical Generalization to parents.}
    \label{fig:editing_details}
\end{figure}

\textbf{Metrics.} For each edited model:
\begin{itemize}[leftmargin=*]
\item \emph{Edit Success}: accuracy on (A, husband, B').
\item \emph{Logical Generalization (Reverse-relation)}: accuracy on (B', wife, A).
\item \emph{Logical Generalization (B', son/daughter, C/D)}: mean accuracy over (B', son/daughter, C/D).
\item \emph{Logical Generalization (C/D, father, B')}: mean accuracy over (C/D, father, B').
\item \emph{Locality (In-Family)}: accuracy on other facts inside the edited family excluding any incident to B'. 
\item \emph{Locality (Other Families)}: accuracy on a fixed held-out set of facts from untouched families.
\end{itemize}
Accuracies are proportion of correct next-token generations for the object name (exact match). Curves in Figure~\ref{fig:editing_details} report the mean over the 50 independent edits.

\end{document}